\documentclass[lettersize,journal]{IEEEtran}
\usepackage{amsmath,amsfonts}
\usepackage{algorithmic}
\usepackage{algorithm}
\usepackage{array}
\usepackage[caption=false,font=normalsize,labelfont=sf,textfont=sf]{subfig}
\usepackage{textcomp}
\usepackage{stfloats}
\usepackage{url}
\usepackage{verbatim}
\usepackage{graphicx}
\usepackage{cite}
\usepackage{amsmath,epstopdf,amssymb,color,url,multirow,verbatim,threeparttable,diagbox,xcolor,makecell,booktabs,colortbl}
\usepackage[numbers,sort&compress]{natbib} 
\definecolor{light-gray}{gray}{0.82}

\newcommand{\tft}{\textbf}

\newcommand{\etal}{\textit{et al.}}
\newcommand{\ccn}[1]{\textcolor{blue}{#1}}
\newcommand{\tb}[1]{\textbf{#1}}
\newcommand{\cred}{\color{red}}
\newcommand{\cblue}{\color{blue}}

\newcommand{\ccb}{\textcolor{blue}}

\usepackage{hyperref}
\begin{document}

\title{Data-Independent Operator: A Training-Free Artifact Representation Extractor for Generalizable Deepfake Detection}

\author{Chuangchuang Tan$^{1,2}$, Ping Liu$^{3}$, RenShuai Tao$^{1,2}$, Huan Liu$^{1,2}$, Yao Zhao$^{1,2}$, Baoyuan Wu$^{4}$, Yunchao Wei$^{1,2}$ \\
{\small $^1$Institute of Information Science, Beijing Jiaotong University, China}\\
{\small $^2$Beijing Key Laboratory of Advanced Information Science and Network Technology, China} \\
{\small $^3$CSE department, University of Nevada, Reno, USA}\\
{\small $^4$School of Data Science, The Chinese University of Hong Kong, Shenzhen (CUHK-Shenzhen), China}

\thanks{Yao Zhao is the Corresponding author}
\thanks{The email of the first author is tanchuangchuang@bjtu.edu.cn}
}


\markboth{Journal of \LaTeX\ Class Files,~Vol.~14, No.~8, August~2021}%
{Shell \MakeLowercase{\textit{et al.}}: A Sample Article Using IEEEtran.cls for IEEE Journals}


\maketitle

\begin{abstract}
Recently, the proliferation of increasingly realistic synthetic images generated by various generative adversarial networks has increased the risk of misuse. 
Consequently, there is a pressing need to develop a generalizable detector for accurately recognizing fake images. 
The conventional methods rely on generating diverse training sources or large pretrained models. 
In this work, we show that, on the contrary, the small and training-free filter is sufficient to capture more general artifact representations. 
Due to its unbias towards both the training and test sources, we define it as Data-Independent Operator (DIO) to achieve appealing improvements on unseen sources. 
In our framework, handcrafted filters and the randomly-initialized convolutional layer can be used as the training-free artifact representations extractor with excellent results.
With the data-independent operator of a popular classifier, such as Resnet50, one could already reach a new state-of-the-art without bells and whistles. 
We evaluate the effectiveness of the DIO on 33 generation models, even DALLE and Midjourney. 
Our detector achieves a remarkable improvement of $13.3\%$, establishing  a new state-of-the-art performance. The DIO and its extension can serve as strong baselines for future methods. The code is available at \url{https://github.com/chuangchuangtan/Data-Independent-Operator}.
\end{abstract}


\section{Introduction}
\IEEEPARstart{R}{ecent} advancements in generation tecnology have significantly improved the generation of realistic synthetic images, like Generative Adversarial Networks (GANs) \cite{goodfellow2014generative, karras2018progressive, karras2019style} and Diffusion \cite{ho2020denoising, rombach2022high}. 
Unfortunately, this progress has also given rise to a surge in the production of forgeries that are nearly indistinguishable from genuine images to the human eye. 
The potential misuse of forged media poses serious risks to society.

To combat this growing concern, researchers have developed various deepfake detectors \cite{wang2020cnn, Frank, frank2020leveraging, jeong2022frepgan, jeong2022fingerprintnet}, some of which are specifically tailored for face forgery detection. 
Nevertheless, most existing forgery detection methods are confined to the same domain during both training and evaluation, limiting their ability to generalize to unseen domains, encompassing unknown generation models or categories. 
This lack of generalization hinders their effectiveness in real-world scenarios where unseen forgery sources are prevalent. 

\begin{figure}[t]
 \centering
  \includegraphics[scale=0.30]{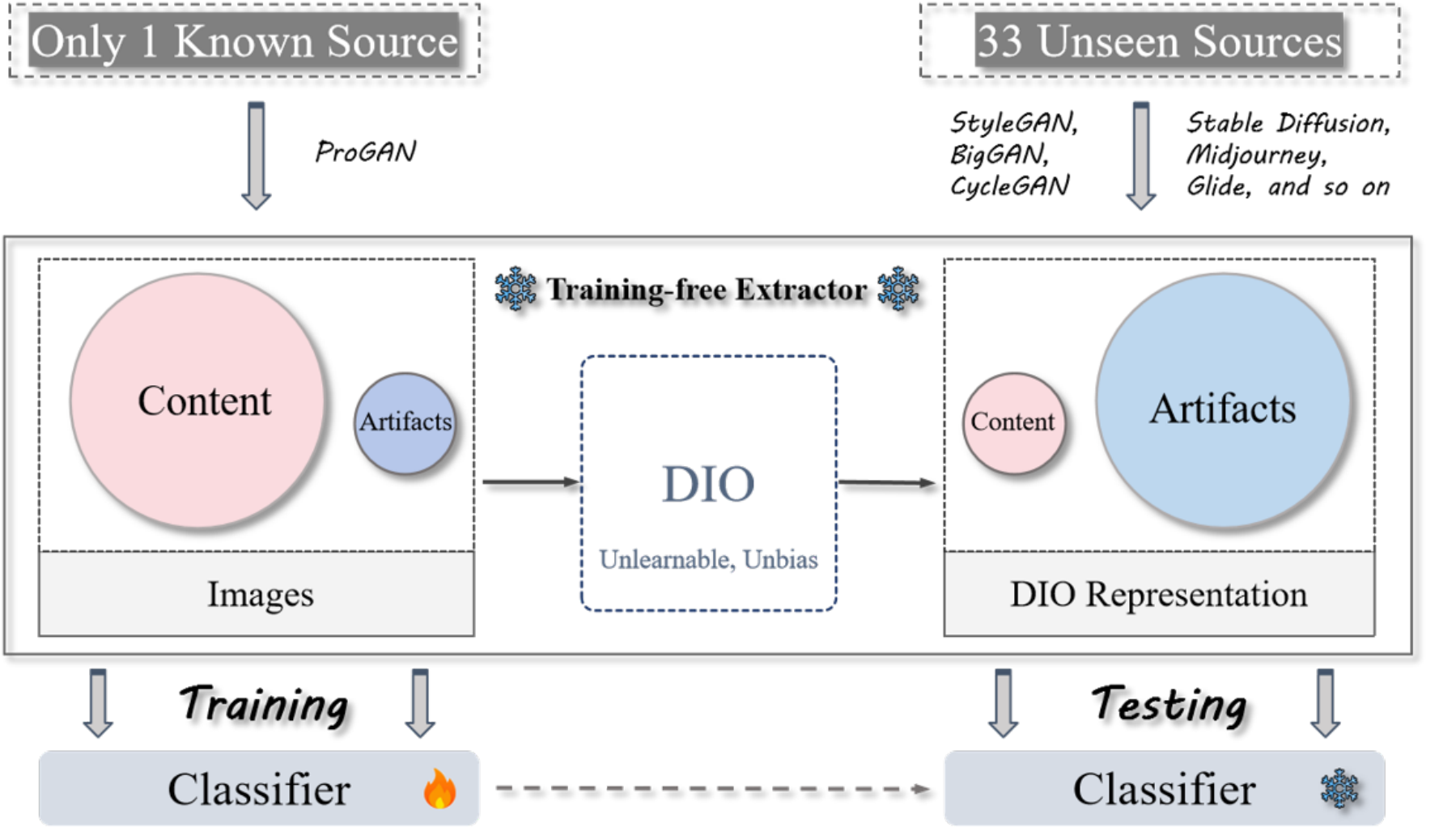}
  \caption{
  In order to enhance the generalization ability of detectors, we employ data-independent operators (DIO) - fixed filters that are unbiased treatment of both known and unseen data - for extracting more general artifact representations. Our DIO effectively suppresses the content of images, compelling the detectors to focus on identifying artifact clues for detecting forged images. Through the incorporation of these filters, the detector trained on a single source demonstrates enhanced performance when tested on 33 previously unseen sources.
  }
  \label{fig:fig1}
\end{figure}

In the realm of generalizable forgery image detection, the primary objective is to discern forged images originating from diverse and unseen generation models, as opposed to merely performing object classification tasks on the images.  
A significant challenge in this context lies in the occurrence of distribution drift between images sourced from different origins. This drift is influenced by both content and artifact factors. 
The content of images may exhibit considerable variability across various sources, encompassing variations in object shapes, structures, and visual attributes. 
Additionally, the distinct artifacts introduced by different Generative Adversarial Network models further contribute to the disparities between source domains. 
Consequently, the detector may become susceptible to overfitting to specific patterns present in the training images, leading to suboptimal performance on unseen sources.
In forgery image detection, the detector should not be dependent on the content (cat) and artifacts (ProGAN) of a specific source but on a more generalizable artifacts representation.
The lack of generalization to these diverse sources poses a fundamental issue.

To address this challenge, the fundamental solution is to develop a source-invariant representation that empowers the detector with the ability to generalize across unseen sources. 
Existing methods \cite{wang2020cnn, jeong2022frepgan, jeong2022fingerprintnet, chen2022ost, ojha2023towards, Tan2023CVPR} often resort to generating representations for all possible sources or employing large pretrained models to train detector, leading to practical constraints and high computational costs.

In light of these limitations, there is a pressing need to devise a simple yet effective source-invariant artifact representation extractor that facilitates generalizable forgery image detection. 
The ideal representation extractor should possess the following characteristics: 1) minimal dependence on training sources, 2) alignment of artifacts {from} diverse domains {to a common space}, and 3) unbiased treatment of both known and unknown data.

In this study, we present a compelling alternative to the conventional artifact representation extractor, demonstrating that it need not solely rely on learning from training data. 
Instead, we propose the novel and straightforward Data-Independent Operator (DIO) framework, which explores the efficacy of fixed filters, encompassing both handcrafted filters and the convolutional layer, to bolster the generalization capabilities of fake image detection. 
The key innovation of our framework lies in leveraging fixed filters that remain independent of the training source, hence aptly named the Data-Independent Operator.  
By adopting this approach, we seek to suppress the content of image and mitigate the discrepancy between different sources in a specific feature space, thereby facilitating the extraction of source-invariant representations and leading to significantly improved detection performance.

Specifically, within the DIO framework, we leverage both handcrafted filters and the pretrained convolutional layer as the data-independent operator to enhance the generalization of fake image detection. 
\ccn{}This combined approach effectively suppress the content of images to force the detectors to mine other clues for forgery image detection. 
Meanwhile, it transforms the distinctive artifacts produced by various generation models into a more generalized representation, achieving alignment across diverse domains in a specific feature space, as shown in Fig. \ref{fig:fig2}b. 
Thanks to its independence from various sources, the DIO exhibits {a less unbiased } nature for both training and unseen sources. This characteristic enables the DIO to effectively capture and represent artifact patterns across diverse datasets and generation models without being influenced by source-specific biases.
In addition, expanding upon the concept introduced by Ulyanov et al. \cite{ulyanov2018deep}, we explore the effectiveness of randomly-initialized neural networks as additional handcrafted priors for the DIO. 
This exploration allows us to further investigate the potential of random filters in extracting a generalized representation.

To augment the capabilities of the DIO, we introduce the Multi-DIOs (MDIO) approach, which combines different data-independent operators in a cascade manner. 
By sequentially applying multiple patterns of DIOs to filter the images, we achieve enhanced artifact representations that are more adaptable and transferable to previously unseen data. 

To comprehensively evaluate the generalization ability of our proposed DIO framework, we conduct simulations using five deepfake detection datasets, including  
 33 distinct models~\footnote{ProGAN, StyleGAN, StyleGAN2, BigGAN, CycleGAN, StarGAN, GauGAN, Deepfake, WFIR, AttGAN, BEGAN, CramerGAN, InfoMaxGAN, MMDGAN, RelGAN, S3GAN, SNGAN, STGAN, ADM, DDPM, IDDPM, LDM, PNDM, VQ-Diffusion, Stable-Diffusion v1, Stable-Diffusion v2, DALLE, Glide, Midjourney, Stable-Diffusion v1.4, Stable-Diffusion v1.5, Wukong, DALLE-2.}.
Our extensive experiments demonstrate the effectiveness and versatility of the artifact representation generated by the DIO across diverse, unseen sources.

Our paper makes the following contributions:
\begin{itemize}
\item We introduce a novel and effective framework, called Data-Independent Operator (DIO), which leverages fixed filters as the artifact extractor to enhance the generalization ability of forgery image detection. 
By using fixed and unlearnable filters as the artifact extractor, our DIO framework reduces the dependence on training data sources and improves detection performance across unseen domains.
\item We extend our approach by introducing the Multiple Data-Independent Operator (MDIO) method. 
This extension combines the artifact representations of different DIOs, enabling enhanced detection performance by leveraging diverse filters for generating more general representations.
\item Our experiments demonstrate that the artifact representation generated by the data-independent operators exhibits strong generalization capabilities across 33 different generation models used for forgery image synthesis. 
This showcases the versatility and robustness of our proposed DIO framework in handling various unseen sources.
\item Through evaluations of wild scenes, our proposed method achieves a new state-of-the-art performance for forgery image detection on unseen sources. We observe a remarkable gain of $13.3\%$ compared to existing methods, highlighting the effectiveness and superiority of our DIO framework in real-world scenarios.
\end{itemize}

\section{Related Work}
In this section, we provide a brief survey on GAN image detection approaches, which can be classified into two main categories: image-based and frequency-based detection.
\subsection{Image-based Fake Detection}
Some previous works \cite{Deepfake} adopt images as the input data to train a binary classification model for forgery detection. 
Rossler \etal \cite{Deepfake} employ images to train a simple Xception \cite{chollet2017xception} for detecting the fake face image. 
Some works focus on the cue on specific regions, such as eyes and lips, to detect fake face media \cite{li2018ictu,haliassos2021lips}. 
Some studies explore the artifacts of the inconsistencies within the forged images \cite{ye2007detecting,zhao2021learning,dong2022protecting}. 
Chai \etal \cite{chai2020makes} use limited receptive fields to find the patches which make images detectable. 
Some works generalize the detectors to unseen sources by enriching the diversity of training data, such as augmentation methods\cite{wang2020cnn, wang2021representative}, adversarial training\cite{chen2022self}, reconstruction\cite{cao2022end, he2021beyond}, fingerprint generator\cite{jeong2022fingerprintnet}, and blending images\cite{shiohara2022detecting}. 
In addition, OST \cite{chen2022ost} uses the re-synthesized test images to enable  one-shot training of the detector during testing, facilitating learning from unknown data. 
Ju \etal \cite{ju2022fusing} fuse the global spatial information and local informative features to train a two-branch model. 
UnivFD \etal \cite{ojha2023towards} and Tan  \etal \cite{Tan2023CVPR} employ the feature map and gradients as  the general representation, respectively.

\subsection{Frequency-based Fake Detection}
As GAN architectures rely on up-scaling operations, there are distinct frequency patterns in the generated images \cite{Durall}. Many detectors 
focus on the artifacts in the frequency spectra. 
LOG \cite{masi2020two} fuses information from the color and frequency domains to detect  manipulated face images and videos. 
\(F^3\)-Net \cite {qian2020thinking} introduces frequency components partition and discrepancy of frequency statistics between real and forged images  into the face forgery detection. 
Luo \etal \cite{luo2021generalizing} utilize multiple high-frequency features of images to improve the generalization performance. 
ADD \cite{woo2022add}  develops two distillation modules  to detect highly compressed deepfakes, including frequency attention distillation and multi-view attention distillation. 
BiHPF \cite{jeong2022bihpf} amplifies the magnitudes of the artifacts by two high-pass filters.  
FreGAN \cite{jeong2022frepgan} finds that unique frequency-level artifacts in generated images can cause overfitting to the training sources, after analyzing frequency artifacts of various GAN models or object categories. Thus, FreGAN tends to reduce the effect of the frequency-level artifacts by the frequency-level perturbation maps.

\section{Methodology}
In this section, we present our proposed framework, Data-Independent Operator (DIO), which is designed to acquire a general artifact representation through fixed filters. 
To address the challenges in generalizable forgery image detection, we commence with a comprehensive review of the topic. 
Subsequently, we define the data-independent operator and outline the overall architecture of our proposed framework. 
Additionally, we introduce Multi-DIOs, an extension of our approach that combines multiple data-independent operators, resulting in enhanced detection performance.

\subsection{Generalizable Forgery Image Detections}
In this section, we embark on a comprehensive review of the task concerning generalizable forgery image detections. 
Throughout our analysis, we address a scenario wherein images from \(n\) diverse sources necessitate detection in real-world settings. 
Consequently, we represent the image dataset sampled from these sources as:
\begin{equation}
\begin{split}
  I = \{  &\{ I_{j}^{s_1}, y_{j}\}_{j=1}^{N_{1}},    \{ I_{j}^{s_2}, y_{j}\}_{j=1}^{N_{2}}, ...  ,\\ & \{ I_{j}^{s_i}, y_{j}\}_{j=1}^{N_{i}},  
     ...  ,\{ I_{j}^{s_n}, y_{j}\}_{j=1}^{N_{n}} \},       \\
 \end{split}
  \label{eq:eq1}
\end{equation}
where \(s_i\) is \(source_{i}\), \(N_i\) is the amount of images from \(source_{i}\), \(I_{j}^{s_i}\) is the \(j\)th image of \(source_{i}\). Every image is labeled with \(y\) as either \(real (y=0)\) or \(fake (y=1)\). In \(source_{i}\), the fake images are generated by one generator, and the real images are sampled from the dataset used to train the generator. 

In real-world scenarios, test images often originate from unknown sources or generator models, introducing contents and artifacts that differ from those present in the training data. Now, we adopt images from \(source_{i}\) to train a binary classifier \(D(\cdot)\), and images from \(source_{j, j \neq i}\) for testing, as described below:
\begin{equation}
 D^{s_{i}} = \mathop{\arg\min}_{\theta} \ L (D( I^{s_i}; \theta),\  y),
  \label{eq:eq3}
\end{equation}
where \(\theta\) is the parameters of detector \(D^{s_{i}}\), \(L()\) is the standard cross-entropy loss. 
The detector \(D^{s_{i}}\) performs the detection on the image dataset \(I\), which includes samples from both  known \(source_{i}\) and unknown \(source_{j, j \neq i}\). 
To develop a generic and effective artifact representation, we emphasize the importance of following characteristics:
\begin{itemize}
\item Suppressing the impact of image content \(C\) on the detector  \(D^{s_{i}}\) to prevent overfitting to training sources.
\item Aligning unique artifacts \(A\) across diverse domains to reduce the distance between them.
\item Ensuring that the extractor remains independent of data to provide unbiased treatment for both  \(source_{i}\) and unknown \(source_{j, j \neq i}\) data.
\
\end{itemize}

\subsection{Data-Independent Operator}

To achieve the aforementioned purposes above, we propose the Data-Independent Operator (DIO) with several key features. First, the DIO includes filters that can effectively suppress the content of the image. Second, the DIO includes filters that can reduce distribution drift between images sampled from different categories or generated by different models. Finally, it is essential that the filters used in the DIO should be unlearnable and do not involve the learned knowledge from the training data. 
We employ various filters to implement data-independent operators, including high-pass filter, low-pass filter, pre-trained convolutional layer, and randomly initialized convolutional layer. 
By applying these filters to the images, their content is suppressed, and they are transformed into a high-dimensional feature space to reduce the unique information, resulting in a reduction of unique information associated with various generation models.  
It is worth noting that these filters are unlearnable during the training phase, which effectively reduces the risk of overfitting.

Based on the proposed data-independent operator, we develop the DIO framework, which integrates this operator and popular classifier to achieve generalizable forgery detection. 
We firstly feed the images \(I^{s_i}\) into data-independent operator to extract the artifact representation \(P\) as:
\begin{equation}
\begin{split}
 P_{DIO} = f^{DIO}(I^{s_i}),
\end{split}
  \label{eq:eq4}
\end{equation}
where the artifact representation $P_{DIO}$ contains the suppressed content and mapped artifacts by DIO.
Then, we use the output of DIO as the artifact representation to train a classifier \(D\): 

\begin{equation}
 D^{s_{i}} = \mathop{\arg\min}_{\theta} \ L (D( P_{DIO}; \theta),\  y).
  \label{eq:eq5}
\end{equation}
The deliberate suppressing of content in the DIO framework is aimed at reducing its informativeness for the detectors $D^{s_i}$ trained on source $s_i$.
This intentional reduction in content information prompts the detectors to explore alternative clues and place greater reliance on the artifact representation for effective forgery image detection.
By emphasizing the importance of $A^{s{i}}_{DIO}$, which encompasses more general and invariant characteristics across diverse domains, the DIO enhances the detectors' ability to detect fake images with improved generalization performance. 


Through the artifact representation obtained by DIO, the classifier focuses on the source-invariant features
and improves detection performance on unseen sources. Extensive experiments on $33$ generation model demonstrate the effectiveness of the data-independent operator. 
Fig. \ref{fig:fig3} also provides the visualization evidence for our explanation. 
Our DIO significantly reduces the differences between images from different sources. 
%

\begin{figure}[t]
  \centering
   \includegraphics[scale=1.2]{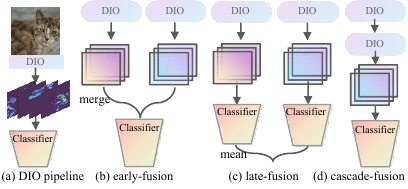}
   \caption{The pipeline of the simple and effective DIO and MDIO. In order to enhance the generalization ability of detectors, we utilize data-independent operators - fixed filters that are unbiased treatments of both known and unseen data - to extract more general artifact representations. With the data-independent operator of a classifier, one could already reach a new SOTA without bells and whistles.}
   \label{fig:fig4}
\end{figure}

\subsection{Multiple Data-Independent Operators}
In order to further improve the generalization of our DIO framework, we propose an advanced Multiple Data-Independent Operator (MDIO) to fuse various data-independent operators, through which multi-patters of DIO can be learned simultaneously.
To investigate the effectiveness of MDIO, we experiment with three different fusion structures, including early-fusion, late-fusion, and cascade-fusion, as shown in Fig. \ref{fig:fig4}. 
{In the early-fusion structure, the artifact representations obtained by different DIOs are merged and fed into the classifier. In the late-fusion structure, we first train classifiers using different data-independent operators, and then average the outputs of the classifiers to obtain the final prediction. In the cascade-fusion structure, multiple data-independent operators are cascaded together to extract the final artifact representation.}

\section{Experimental Results}
\label{ER}
\subsection{Dataset}
\subsubsection{Training set.} For a fair comparison, we utilize the training set from ForenSynths\cite{wang2020cnn} to train our detectors, following the baseline approaches\cite{wang2020cnn,jeong2022bihpf,jeong2022frepgan}. 
The training set comprises $20$ categories, each containing $18,000$ synthetic images generated by ProGAN, and an equal number of real images sourced from the LSUN dataset. 
In line with previous works\cite{jeong2022bihpf,jeong2022frepgan}, we adopt the $(car, cat, chair, horse)$ configurations as our 4-class training settings, respectively.

\subsubsection{Test set.} To comprehensively assess the generalization capability of our proposed approach, we construct a diverse test set that includes a wide range of images and GAN/Diffusion models. 

$\bullet$ ForenSynths\cite{wang2020cnn}: The test set includes fake images generated by 8 generation models~\footnote{ProGAN\cite{karras2018progressive}, StyleGAN\cite{karras2019style}, StyleGAN2\cite{karras2020analyzing}, BigGAN\cite{BigGAN}, CycleGAN\cite{CycleGAN}, StarGAN\cite{choi2018stargan}, GauGAN\cite{GauGAN} and Deepfake\cite{Deepfake}.}. Real images are sampled from 6 datasets (LSUN\cite{yu2015lsun}, ImageNet\cite{russakovsky2015imagenet}, CelebA\cite{CelebA}, CelebA-HQ\cite{karras2018progressive}, COCO\cite{coco}, and FaceForensics++\cite{Deepfake}).

$\bullet$ GANGen-Detection\cite{chuangchuangtan-GANGen-Detection}: To replicate the unpredictability of wild scenes, we extend our evaluation by collecting images generated by 9 additional GANs~\footnote{AttGAN\cite{AttGAN}, BEGAN\cite{began}, CramerGAN\cite{CramerGAN}, InfoMaxGAN\cite{InfoMaxGAN}, MMDGAN\cite{MMDGAN}, RelGAN\cite{RelGAN}, S3GAN\cite{S3GAN}, SNGAN\cite{SNGAN}, and STGAN\cite{STGAN}}. 
There are 4K test images for each model, with equal numbers of real and fake images.

$\bullet$ DiffusionForensics\cite{Wang_2023_ICCV}: To expand the testing scope, we adopt the diffusions dataset of DIRE \cite{Wang_2023_ICCV} for evaluation, including ADM\cite{dhariwal2021diffusion}, DDPM\cite{ho2020denoising}, IDDPM\cite{nichol2021improved}, LDM\cite{rombach2022high}, PNDM\cite{liu2022pseudo}, Vqdiffusion\cite{gu2022vector}, Stable Diffusion v1\cite{rombach2022high}, Stable Diffusion v2\cite{rombach2022high}. 
The real images are sampled from LSUN\cite{yu2015lsun} and ImageNet\cite{russakovsky2015imagenet} datasets.

$\bullet$ UniversalFakeDetect\cite{ojha2023towards}:  This test set contains images generated from ADM \cite{dhariwal2021diffusion}, Glide\cite{nichol2021glide}, DALL-E-mini\cite{ramesh2021zero}, LDM\cite{rombach2022high}. 
It adopts images of  LAION\cite{schuhmann2021laion} and ImageNet\cite{russakovsky2015imagenet} datasets as the real data.

$\bullet$ AIGCDetectBenchmark\cite{rptc}: This test set collects 7 GAN-based generative models and whichfaceisreal (WFIR) from ForenSynths\cite{wang2020cnn}, and collects 7 Diffusion models~\footnote{ADM\cite{dhariwal2021diffusion}, Glide\cite{nichol2021glide}, Midjourney\cite{Midjourney}, SDv1.4\cite{rombach2022high}, SDv1.5\cite{rombach2022high}, VQDM\cite{gu2022vector}, Wukong\cite{Wukong}} from GenImage\cite{zhu2023genimage}, then generate images from DALLE 2 \cite{ramesh2022hierarchical}.


\subsection{Implementation Details}

We adopt the ResNet50\cite{he2016deep} model, which is pre-trained with ImageNet\cite{russakovsky2015imagenet}, as the classifier for our experiments. 
The ResNet50 model is trained using the Adam optimizer\cite{kingma2015adam} with a learning rate of \(2 \times 10^{-4}\). 
To accommodate different DIOs, we modify the number of channels in the first convolutional layer of ResNet50. 
For the DIO framework, we set the batch size to 128 and train for 100 epochs. 
As for the Multi-DIOs (MDIO), the batch size is set to 32, and training is performed for 40 epochs. 
We apply a learning rate decay strategy, reducing the learning rate by ten percent after every ten epochs for DIO and four epochs for MDIO, respectively. To assess the performance of the proposed method, we follow the evaluation metrics used in the baselines\cite{jeong2022bihpf,jeong2022frepgan}, which include the average precision score (A.P.) and accuracy (Acc.). 
Our method is implemented using the PyTorch on Nvidia GeForce RTX A4000 GPU.

\begin{table*}[!ht]
  \caption{
  {Effect of different data-independent operators.} We show the performance of Resnet50\cite{he2016deep} on unseen sources with and without data-independent operators.
}
\centering
\resizebox{\textwidth}{20.3mm}{
    \begin{tabular}{l  c c c c c c c c c c c c c c c c| c c}
    \bottomrule \hline
       \multirow{2}*{Diff. DIOs} & \multicolumn{2}{c}{ProGAN}& \multicolumn{2}{c}{StyleGAN}& \multicolumn{2}{c}{StyleGAN2}& \multicolumn{2}{c}{BigGAN}& \multicolumn{2}{c}{CycleGAN}& \multicolumn{2}{c}{StarGAN}& \multicolumn{2}{c}{GauGAN}& \multicolumn{2}{c|}{Deepfake}& \multicolumn{2}{c}{Mean}\\
         \cline{2-19} ~   & Acc. & A.P. & Acc. & A.P. & Acc. & A.P. & Acc. & A.P. & Acc. & A.P. & Acc. & A.P. & Acc. & A.P. & Acc. & A.P. & Acc. & A.P.\\  \bottomrule \hline
w/o DIO                                                     & 99.8 & 100.0 & 82.6 & 90.3 & 74.9 & 98.4 & 59.8 & 63.4 & 63.5 & 71.1 & 100.0 & 100.0 & 55.6 & 55.2 & 73.4 & 95.7 & 76.2 & 84.3 \\
DIO-LoG                                                    & 99.7 & 100.0 & 84.2 & 94.9 & 82.7 & 98.9 & 65.3 & 65.9 & 72.4 & 83.9 & 100.0 & 100.0 & 65.4 & 67.0 & 76.4 & 95.4 & 80.8(\tb{+4.6}) & 88.3\\
DIO-Laplacian                                             & 99.8 & 100.0 & 85.5 & 97.7 & 85.9 & 99.6 & 69.1 & 69.3 & 78.0 & 83.7 & 100.0 & 100.0 & 61.8 & 56.9 & 73.3 & 86.2 & 81.7(\tb{+5.5}) & 86.7 \\
DIO-Avgpool                                              & 99.5 & 100.0 & 83.1 & 92.4 & 84.7 & 99.1 & 67.7 & 64.1 & 71.0 & 75.5 & 100.0 & 100.0 & 59.1 & 59.1 & 90.4 & 95.7 & 81.9(\tb{+5.7}) & 85.7 \\
DIO-Sobel                                                  & 97.4 & 100.0 & 88.5 & 99.5 & 86.8 & 99.6 & 76.2 & 78.3 & 78.2 & 84.1 & 99.7 & 100.0 & 62.0 & 59.6 & 67.6 & 92.3 & 82.0(\tb{+5.8}) & 89.2 \\
 DIO-VGG16 \nocite{2014Very}                    & 99.5 & 100.0 & 86.4 & 97.0 & 97.9 & 100.0 & 75.8 & 77.4 & 79.4 & 88.5 & 100.0 & 100.0 & 61.9 & 59.9 & 57.9 & 87.9 & 82.3(\tb{+6.1}) & 88.8 \\
 DIO-Res.101 \nocite{he2016deep}              & 99.6 & 100.0 & 88.8 & 98.8  & 93.1 & 99.9 & 76.9 & 78.7 & 80.1 & 87.4 & 99.7 & 100.0 & 63.7 & 64.0 & 72.2 & 95.6 & 84.3(\tb{+8.1}) & 90.5  \\
DIO-Incep. \nocite{szegedy2016rethinking}   & 99.7 & 100.0 & 88.5 & 98.3 & 97.9 & 100.0 & 74.2 & 78.2 & 78.7 & 89.7 & 99.9 & 100.0 & 62.1 & 58.7 & 72.1 & 89.5 & 84.1(\tb{+7.9}) & 89.3 \\
DIO-CLIP \nocite{CLIP}                               & 99.7 & 100.0 & 83.5 & 93.0 & 87.4 & 99.7 & 72.2 & 74.0 & 78.6 & 87.4 & 100.0 & 100.0 & 54.6 & 48.9 & 68.8 & 85.7 & 80.6(\tb{+4.4}) & 86.1 \\
DIO-Deep.v3 \nocite{DeeplabV3}                & 99.4 & 100.0 & 90.7 & 98.8 & 94.9 & 99.9 & 73.3 & 74.1 & 79.5 & 85.5 & 99.2 & 100.0 & 64.3 & 62.9 & 60.2 & 87.0 & 82.7(\tb{+6.5}) & 88.5 \\
 \bottomrule
    \end{tabular}
}

  \label{tab:tab1}
\end{table*}

\begin{table*}[!ht]
  \caption{
  {Effect of kernel shape of Randomly Initialized Convolutional Layers.}  We employ Randomly Initialized Convolutional Layers as the artifact extractor to prove that prior knowledge is unnecessary for the data-independent operator.}
    \centering
\resizebox{\textwidth}{30mm}{
    \begin{tabular}{c c c c c c c c c c c c c c c c c| c c}
    \bottomrule   \hline
     \multirow{2}*{\makecell[c]{kernel shape}}  & \multicolumn{2}{c}{ProGAN}& \multicolumn{2}{c}{StyleGAN}& \multicolumn{2}{c}{StyleGAN2}& \multicolumn{2}{c}{BigGAN}& \multicolumn{2}{c}{CycleGAN}& \multicolumn{2}{c}{StarGAN}&        \multicolumn{2}{c}{GauGAN}& \multicolumn{2}{c|}{Deepfake}& \multicolumn{2}{c}{Mean}\\ 
      \cline{2-19} ~   & Acc. & A.P. & Acc. & A.P. & Acc. & A.P. & Acc. & A.P. & Acc. & A.P. & Acc. & A.P. & Acc. & A.P. & Acc. & A.P. & Acc. & A.P.\\ \bottomrule \hline
(3, 8, 1, 1) & 99.5 & 100.0 & 91.2 & 99.4 & 96.6 & 99.9 & 75.1 & 75.8 & 80.7 & 89.7 & 99.7 & 100.0 & 67.8 & 68.6 & 69.2 & 89.7 & 85.0 & 90.4 \\
(3, 8, 3, 3)  & 98.7 & 99.9 & 80.7 & 95.9 & 88.1 & 99.1 & 59.8 & 67.6 & 77.6 & 86.4 & 98.4 & 100.0 & 63.7 & 69.6 & 55.0 & 80.2 & 77.7 & 87.3 \\
(3, 8, 5, 5) & 98.3 & 99.9 & 74.7 & 95.5 & 82.1 & 99.2 & 57.5 & 70.9 & 71.5 & 85.0 & 85.6 & 99.9 & 63.1 & 76.8 & 54.1 & 88.1 & 73.4 & 89.4 \\
\hline
(3, 16, 1, 1) & 99.4 & 100.0 & 90.4 & 98.8 & 97.7 & 100.0 & 73.8 & 75.2 & 83.5 & 91.8 & 99.4 & 100.0 & 71.8 & 73.0 & 70.1 & 84.3 & 85.8 & 90.4 \\
(3, 16, 3, 3) & 98.8 & 100.0 & 86.0 & 97.8 & 92.8 & 99.9 & 67.8 & 75.1 & 80.8 & 89.2 & 98.2 & 100.0 & 59.7 & 64.4 & 67.0 & 94.3 & 81.4 & 90.1 \\
(3, 16, 5, 5) & 98.5 & 100.0 & 87.6 & 98.4 & 93.6 & 99.8 & 67.6 & 74.4 & 76.7 & 85.7 & 92.5 & 99.9 & 70.5 & 73.5 & 79.1 & 88.6 & 83.2 & 90.0 \\
\hline
(3, 32, 1, 1) & 99.2 & 100.0 & 89.9 & 99.1 & 97.2 & 100.0 & 73.7 & 75.0 & 82.4 & 90.2 & 99.8 & 100.0 & 67.7 & 68.4 & 79.7 & 89.7 & 86.2 & 90.3 \\
(3, 32, 3, 3) & 98.8 & 100.0 & 89.2 & 98.9 & 94.4 & 100.0 & 69.5 & 74.7 & 83.1 & 90.3 & 97.1 & 100.0 & 69.6 & 73.2 & 71.5 & 91.4 & 84.2 & \tft{91.0} \\
(3, 32, 5, 5) & 98.9 & 100.0 & 89.3 & 98.9 & 95.5 & 99.9 & 68.4 & 75.0 & 82.9 & 91.4 & 95.3 & 100.0 & 69.8 & 72.6 & 66.5 & 88.3 & 83.3 & 90.8 \\
\hline
(3, 64, 1, 1) & 99.1 & 100.0 & 91.1 & 99.1 & 97.7 & 100.0 & 73.3 & 72.4 & 79.9 & 86.7 & 99.7 & 100.0 & 67.9 & 67.0 & 82.6 & 92.1 & \tft{86.4} & 89.7 \\
(3, 64, 3, 3) & 99.2 & 100.0 & 90.9 & 99.4 & 97.2 & 100.0 & 75.1 & 75.5 & 81.3 & 87.9 & 99.9 & 100.0 & 67.5 & 66.0 & 59.4 & 91.9 & 83.8 & 90.1 \\
(3, 64, 5, 5) & 98.9 & 99.9 & 87.6 & 98.3 & 93.5 & 99.9 & 70.0 & 74.9 & 76.4 & 84.4 & 84.2 & 100.0 & 68.2 & 69.7 & 54.0 & 85.2 & 79.1 & 89.1 \\

\hline

(3, 128, 1, 1) & 99.5 & 100.0 & 90.9 & 99.3 & 97.6 & 100.0 & 75.6 & 74.6 & 80.7 & 87.0 & 95.5 & 100.0 & 66.8 & 67.9 & 61.9 & 88.0 & 83.5 & 89.6 \\
(3, 128, 3, 3) & 98.7 & 99.9 & 90.4 & 99.4 & 96.6 & 99.9 & 75.7 & 76.9 & 81.8 & 89.5 & 99.7 & 100.0 & 68.8 & 70.5 & 63.8 & 89.0 & 84.4 & 90.6 \\
(3, 128, 5, 5) & 99.0 & 100.0 & 91.2 & 99.6 & 97.1 & 100.0 & 74.2 & 75.2 & 80.8 & 87.0 & 99.0 & 100.0 & 70.4 & 71.3 & 64.1 & 89.5 & 84.5 & 90.3 \\
 \bottomrule 
    \end{tabular}
}

  \label{tab:tab2}
\end{table*}

\begin{table*}[!ht]
    \centering
      \caption{Effect of fusion structures. We show cross-model performance of Multiple DIOs with different fusion structures on the test set of ForenSynths\cite{wang2020cnn}.}
\resizebox{\textwidth}{9mm}{
    \begin{tabular}{c c c c c c c c c c c c c c c c c c| c c}
    \hline 
      \multirow{2}*{\makecell[c]{Fusion}}  & \multirow{2}*{\makecell[c]{DIOs}}   & \multicolumn{2}{c}{ProGAN}& \multicolumn{2}{c}{StyleGAN}& \multicolumn{2}{c}{StyleGAN2}& \multicolumn{2}{c}{BigGAN}& \multicolumn{2}{c}{CycleGAN}& \multicolumn{2}{c}{StarGAN}&        \multicolumn{2}{c}{GauGAN}& \multicolumn{2}{c|}{Deepfake}& \multicolumn{2}{c}{Mean}\\
      \cline{3-20} ~  & ~ &  Acc. & A.P. & Acc. & A.P. & Acc. & A.P. & Acc. & A.P. & Acc. & A.P. & Acc. & A.P. & Acc. & A.P. & Acc. & A.P. & Acc. & A.P.\\ \hline
Early & Random,Sobel &  99.3 & 100.0 & 89.2 & 98.6 & 94.0 & 99.9 & 71.7 & 76.2 & 84.0 & 92.3 & 99.8 & 100.0 & 66.3 & 67.6 & 77.0 & 91.2 & 85.2 & 90.7 \\
Late & Random,Sobel  & 99.8 & 100.0 & 93.2 & 100.0 & 98.6 & 100.0 & 78.5 & 78.3 & 80.7 & 88.5 & 100.0 & 100.0 & 67.0 & 69.7 & 77.4 & 93.5 & 86.9 & 91.2  \\
Cascade & Random,Sobel & 99.7 & 100.0 & 99.5 & 100.0 & 99.4 & 99.9 & 84.8 & 88.0 & 94.4 & 96.7 & 98.1 & 99.0 & 81.6 & 84.1 & 87.8 & 89.4 & 93.2 & 94.6 \\
%
\hline
    \end{tabular}
    }
  \label{tab:tab3}
\end{table*}

\begin{table*}[!ht]
 \vspace{-0.2 cm}
   \caption{\tft{Cross-GAN-Sources Evaluation on the test set of ForenSynths\cite{wang2020cnn}.} The results of \cite{wang2020cnn,Frank,Durall,jeong2022bihpf,jeong2022frepgan} are from \cite{jeong2022frepgan, jeong2022bihpf}. 
  {\cred{Red}} and {\cblue{Blue}} represent the best and second-best performance, respectively. }
    \centering
\resizebox{\textwidth}{24mm}{
    \begin{tabular}{l  c c c c c c c c c c c c c c c c | c c}
    \bottomrule \hline
       \multirow{2}*{Method} & \multicolumn{2}{c}{ProGAN}& \multicolumn{2}{c}{StyleGAN}& \multicolumn{2}{c}{StyleGAN2}& \multicolumn{2}{c}{BigGAN}& \multicolumn{2}{c}{CycleGAN}& \multicolumn{2}{c}{StarGAN}& \multicolumn{2}{c}{GauGAN}& \multicolumn{2}{c|}{Deepfake}& \multicolumn{2}{c}{Mean}\\

         \cline{2-19} ~   & Acc. & A.P. & Acc. & A.P. & Acc. & A.P. & Acc. & A.P. & Acc. & A.P. & Acc. & A.P. & Acc. & A.P.  & Acc. & A.P.  & Acc. & A.P. \\ \bottomrule \hline
        CNNDet\cite{wang2020cnn}      & 91.4 & 99.4 & 63.8 & 91.4 & 76.4 & 97.5 & 52.9 & 73.3 & 72.7 & 88.6 & 63.8 & 90.8 & 63.9 & 92.2 & 51.7 & 62.3 & 67.1  &86.9 \\ 
        FreDect\cite{Frank}                & 90.3 & 85.2 & 74.5 & 72.0 & 73.1 & 71.4 & 88.7 & 86.0 & 75.5 & 71.2 & 99.5 & 99.5 & 69.2 & 77.4 & 60.7 & 49.1 & 78.9 & 76.5 \\ 
        Durall\cite{Durall}              & 81.1 & 74.4 & 54.4 & 52.6 & 66.8 & 62.0 & 60.1 & 56.3 & 69.0 & 64.0 & 98.1 & 98.1 & 61.9 & 57.4 & 50.2 & 50.0 & 67.7  & 64.4\\ 
        Patchfor\cite{chai2020makes}    & 97.8 & 100.0 & 82.6 & 93.1 & 83.6 & 98.5 & 64.7 & 69.5 & 74.5 & 87.2 & 100.0 & 100.0 & 57.2 & 55.4 & 85.0 & 93.2 & 80.7 & 87.1  \\
        F3Net\cite{qian2020thinking}      & 99.4 & 100.0& 92.6 & 99.7 & 88.0 & 99.8 & 65.3 & 69.9 & 76.4 & 84.3 & 100.0& 100.0& 58.1 & 56.7 & 63.5 & 78.8 & 80.4 & 86.2 \\
        SelfBland\cite{shiohara2022detecting}  & 58.8 & 65.2 & 50.1 & 47.7 & 48.6 & 47.4 & 51.1 & 51.9 & 59.2 & 65.3 & 74.5 & 89.2 & 59.2 & 65.5 & 93.8 & 99.3 & 61.9 & 66.4  \\
        GANDetection\cite{mandelli2022detecting} & 82.7 & 95.1 & 74.4 & 92.9 & 69.9 & 87.9 & 76.3 & 89.9 & 85.2 & 95.5 & 68.8 & 99.7 & 61.4 & 75.8 & 60.0 & 83.9 & 72.3 & 90.1  \\
        BiHPF\cite{jeong2022bihpf}       & 90.7 & 86.2 & 76.9 & 75.1 & 76.2 & 74.7 & 84.9 & 81.7 & 81.9 & 78.9 & 94.4 & 94.4 & 69.5 & 78.1 & 54.4 & 54.6 & 78.6  & 77.9\\
        FrePGAN\cite{jeong2022frepgan}    & 99.0 & 99.9 & 80.7 & 89.6 & 84.1 & 98.6 & 69.2 & 71.1 & 71.1 & 74.4 & 99.9 & 100.0& 60.3 & 71.7 & 70.9 & 91.9 & 79.4  & 87.2\\ 
        LGrad\cite{Tan2023CVPR}         & 99.9 & 100.0& 94.8 & 99.9 & 96.0 & 99.9 & 82.9 & 90.7 & 85.3 & 94.0 & 99.6 & 100.0& 72.4 & 79.3 & 58.0 & 67.9 & 86.1 & 91.5 \\
        UnivFD\cite{ojha2023towards}    & 99.7 & 100.0& 89.0 & 98.7 & 83.9 & 98.4 & 90.5 & 99.1 & 87.9 & 99.8 & 91.4 & 100.0& 89.9 & 100.0& 80.2 & 90.2 & \cblue{{89.1}} & \cred{{98.3}} \\
       DIO                     & 99.1 & 100.0 & 91.1 & 99.1 & 97.7 & 100.0 & 73.3 & 72.4 & 79.9 & 86.7 & 99.7 & 100.0 & 67.9 & 67.0 & 82.6 & 92.1 & 86.4 & 89.7 \\
\rowcolor{light-gray} MDIO                    & 99.7 & 100.0 & 99.5 & 100.0 & 99.4 & 99.9 & 84.8 & 88.0 & 94.4 & 96.7 & 98.1 & 99.0 & 81.6 & 84.1 & 87.8 & 89.4 & \cred{93.2} & \cblue{94.6} \\
\bottomrule
    \end{tabular}
}

  \label{tab:SOTA_ForenSynths}
\end{table*}

\begin{table*}[!ht]
 \vspace{-0.2 cm}
   \caption{\textbf{Cross-GAN-Sources Evaluation on the GANGen-Detection\cite{chuangchuangtan-GANGen-Detection} dataset.}}
    \centering
\resizebox{\textwidth}{21mm}{
    \begin{tabular}{l  c c c c c c c c c c c c c c c c c c| c c}
    \bottomrule \hline
       \multirow{2}*{Method} & \multicolumn{2}{c}{AttGAN}& \multicolumn{2}{c}{BEGAN}& \multicolumn{2}{c}{CramerGAN}& \multicolumn{2}{c}{InfoMaxGAN}& \multicolumn{2}{c}{MMDGAN}& \multicolumn{2}{c}{RelGAN}& \multicolumn{2}{c}{S3GAN}& \multicolumn{2}{c}{SNGAN}&  \multicolumn{2}{c|}{STGAN}& \multicolumn{2}{c}{Mean}\\
         \cline{2-21} ~   & Acc. & A.P. & Acc. & A.P. & Acc. & A.P. & Acc. & A.P. & Acc. & A.P. & Acc. & A.P. & Acc. & A.P. & Acc. & A.P. & Acc. & A.P. & Acc. & A.P.\\ \bottomrule \hline
CNNDet\cite{wang2020cnn}                 & 51.1 & 83.7 & 50.2 & 44.9 & 81.5 & 97.5 & 71.1 & 94.7  & 72.9 & 94.4 & 53.3 & 82.1 & 55.2 & 66.1 & 62.7 & 90.4 & 63.0 & 92.7 & 62.3 & 82.9 \\
FreDect\cite{Frank}                                            & 65.0 & 74.4 & 39.4 & 39.9 & 31.0 & 36.0 & 41.1 & 41.0 & 38.4 & 40.5 & 69.2 & 96.2 & 69.7 & 81.9 & 48.4 & 47.9 & 25.4 & 34.0 & 47.5 & 54.7 \\
Durall\cite{Durall}                                            & 39.9 & 38.2 & 48.2 & 30.9 & 60.9 & 67.2 & 50.1 & 51.7 & 59.5 & 65.5 & 80.0 & 88.2 & 87.3 & 97.0 & 54.8 & 58.9 & 62.1 & 72.5 & 60.3 & 63.3 \\
Patchfor\cite{chai2020makes}                        & 68.0 & 92.9 & 97.1 & 100.0 & 97.8 & 99.9 & 93.6 & 98.2 & 97.9 & 100.0 & 99.6 & 100.0 & 66.8 & 68.1 & 97.6 & 99.8 & 92.7 & 99.8 & \cblue{90.1} & 95.4 \\
F3Net\cite{qian2020thinking}   & 85.2 & 94.8 & 87.1 & 97.5 & 89.5 & 99.8 & 67.1 & 83.1 & 73.7 & 99.6 & 98.8 & 100.0 & 65.4 & 70.0 & 51.6 & 93.6 & 60.3 & 99.9 & 75.4 & 93.1 \\
SelfBland\cite{shiohara2022detecting}           & 63.1 & 66.1 & 56.4 & 59.0 & 75.1 & 82.4 & 79.0 & 82.5 & 68.6 & 74.0 & 73.6 & 77.8 & 53.2 & 53.9 & 61.6 & 65.0 & 61.2 & 66.7 & 65.8 & 69.7 \\
GANDetection\cite{mandelli2022detecting}    & 57.4 & 75.1 & 67.9 & 100.0 & 67.8 & 99.7 & 67.6 & 92.4 & 67.7 & 99.3 & 60.9 & 86.2 & 69.6 & 83.5 & 66.7 & 90.6 & 69.6 & 97.2 & 66.1 & 91.6 \\
LGrad\cite{Tan2023CVPR}   & 68.6 & 93.8 & 69.9 & 89.2 & 50.3 & 54.0 & 71.1 & 82.0 & 57.5 & 67.3 & 89.1 & 99.1 & 78.5 & 86.0 & 78.0 & 87.4 & 54.8 & 68.0 & 68.6 & 80.8\\
UnivFD\cite{ojha2023towards}  & 78.5 & 98.3 & 72.0 & 98.9 & 77.6 & 99.8 & 77.6 & 98.9 & 77.6 & 99.7 & 78.2 & 98.7 & 85.2 & 98.1 & 77.6 & 98.7 & 74.2 & 97.8 & 77.6 & \cred{98.8}\\
DIO    & 84.3 & 93.1 & 81.4 & 99.9 & 96.4 & 99.6 & 96.1 & 99.2 & 96.5 & 99.6 & 92.1 & 99.7 & 65.0 & 69.6 & 95.2 & 98.0 & 96.2 & 100.0 & {89.2} & 95.4\\
\rowcolor{light-gray}MDIO                                       & 99.8 & 100.0 & 92.2 & 93.0 & 95.5 & 95.5 & 93.4 & 94.5 & 95.5 & 95.5 & 100.0 & 100.0 & 87.5 & 92.8 & 94.5 & 94.9 & 98.9 & 99.5 & \cred{95.2} & \cblue{96.2}\\
\bottomrule
    \end{tabular}
}

  \label{tab:SOTA_GANGen}
\end{table*}

\begin{table*}[!ht]
  \caption{\textbf{Cross-Diffusion-Sources Evaluation on the test of DiffusionForensics \cite{Wang_2023_ICCV}.}}
\vspace{-0.2 cm}
    \centering
\resizebox{\textwidth}{22mm}{
    \begin{tabular}{l  c c c c c c c c c c c c c c c c | c c}
    \bottomrule \hline
       \multirow{2}*{Method} & \multicolumn{2}{c}{ADM}& \multicolumn{2}{c}{DDPM}& \multicolumn{2}{c}{IDDPM}& \multicolumn{2}{c}{LDM}& \multicolumn{2}{c}{PNDM}& \multicolumn{2}{c}{VQ-Diffusion} & \multicolumn{2}{c}{\makecell[c]{\small{Stable} \\ \small{Diffusion v1} } }   & \multicolumn{2}{c}{\makecell[c]{\small{Stable} \\ \small{Diffusion v2} } }    & \multicolumn{2}{c}{Mean}\\
         \cline{2-19} ~   & Acc. & A.P. & Acc. & A.P. & Acc. & A.P. & Acc. & A.P. & Acc. & A.P. & Acc. & A.P. & Acc. & A.P.  & Acc. & A.P.  & Acc. & A.P. \\ \bottomrule \hline
        CNNDet\cite{wang2020cnn}                            & 53.9 & 71.8 & 62.7 & 76.6 & 50.2 & 82.7 & 50.4 & 78.7 & 50.8 & 90.3 & 50.0 & 71.0 & 38.0 & 76.7 & 52.0 & 90.3 & 51.0 & 79.8 \\
        FreDect\cite{Frank}                                                       & 58.9 & 65.9 & 37.0 & 27.6 & 51.4 & 65.0 & 51.7 & 48.5 & 44.0 & 38.2 & 51.7 & 66.7 & 32.8 & 52.3 & 40.8 & 37.5 & 46.0 & 50.2 \\
        Durall\cite{Durall}                                                       & 39.8 & 42.1 & 52.9 & 49.8 & 55.3 & 56.7 & 43.1 & 39.9 & 44.5 & 47.3 & 38.6 & 38.3 & 39.5 & 56.3 & 62.1 & 55.8 & 47.0 & 48.3 \\
        Patchfor\cite{chai2020makes}                                   & 77.5 & 93.9 & 62.3 & 97.1 & 50.0 & 91.6 & 99.5 & 100.0 & 50.2 & 99.9 & 100.0 & 100.0 & 90.7 & 99.8 & 94.8 & 100.0 & 78.1 & 97.8\\
        F3Net\cite{qian2020thinking}                                     & 80.9 & 96.9 & 84.7 & 99.4 & 74.7 & 98.9 & 100.0 & 100.0 & 72.8 & 99.5 & 100.0 & 100.0 & 73.4 & 97.2 & 99.8 & 100.0 & 85.8 & \cblue{99.0} \\
        SelfBland\cite{shiohara2022detecting}                      & 57.0 & 59.0 & 61.9 & 49.6 & 63.2 & 66.9 & 83.3 & 92.2 & 48.2 & 48.2 & 77.2 & 82.7 & 46.2 & 68.0 & 71.2 & 73.9 & 63.5 & 67.6 \\
        GANDetection\cite{mandelli2022detecting}              & 51.1 & 53.1 & 62.3 & 46.4 & 50.2 & 63.0 & 51.6 & 48.1 & 50.6 & 79.0 & 51.1 & 51.2 & 39.8 & 65.6 & 50.1 & 36.9 & 50.8 & 55.4 \\
        LGrad\cite{Tan2023CVPR}                                      & 86.4 & 97.5 & 99.9 & 100.0 & 66.1 & 92.8 & 99.7 & 100.0 & 69.5 & 98.5 & 96.2 & 100.0 & 90.4 & 99.4 & 97.1 & 100.0 & \cblue{88.2} & {98.5} \\
        UnivFD\cite{ojha2023towards}                                     & 78.4 & 92.1 & 72.9 & 78.8 & 75.0 & 92.8 & 82.2 & 97.1 & 75.3 & 92.5 & 83.5 & 97.7 & 56.4 & 90.4 & 71.5 & 92.4 & 74.4 & 91.7\\
 \rowcolor{light-gray}MDIO  & 94.8 & 99.2 & 99.8 & 99.8 & 99.3 & 99.9 & 99.8 & 99.9 & 99.4 & 99.9 & 99.8 & 99.9 & 97.1 & 99.7 & 99.8 & 99.9 & \cred{98.7} & \cred{99.8}  \\

\bottomrule
 \end{tabular}}
  \label{tab:iccvdataset}
        \vspace{-0.25 cm}
\end{table*}

\begin{table*}[!ht]
 \vspace{-0.2 cm}
   \caption{\textbf{Cross-Diffusion-Sources Evaluation on the diffusion test set of UnivFD \cite{ojha2023towards} .}}
    \centering
\resizebox{\textwidth}{21mm}{
    \begin{tabular}{l  c c c c c c c c c c c c c c c c | c c}
    \bottomrule \hline
       \multirow{2}*{Method} & \multicolumn{2}{c}{DALLE}& \multicolumn{2}{c}{Glide\_100\_10}& \multicolumn{2}{c}{Glide\_100\_27}& \multicolumn{2}{c}{Glide\_50\_27} & \multicolumn{2}{c}{ADM} & \multicolumn{2}{c}{LDM\_100} & \multicolumn{2}{c}{LDM\_200} & \multicolumn{2}{c}{LDM\_200\_cfg}& \multicolumn{2}{c}{Mean}\\
         \cline{2-19} ~   & Acc. & A.P. & Acc. & A.P. & Acc. & A.P. & Acc. & A.P. & Acc. & A.P. & Acc. & A.P. & Acc. & A.P. & Acc. & A.P. & Acc. & A.P.\\ \bottomrule \hline

CNNDet\cite{wang2020cnn}            & 51.8 & 61.3 & 53.3 & 72.9 & 53.0 & 71.3 & 54.2 & 76.0 & 54.9 & 66.6 & 51.9 & 63.7 & 52.0 & 64.5 & 51.6 & 63.1 & 52.8 & 67.4 \\
FreDect\cite{Frank}                         & 57.0 & 62.5 & 53.6 & 44.3 & 50.4 & 40.8 & 52.0 & 42.3 & 53.4 & 52.5 & 56.6 & 51.3 & 56.4 & 50.9 & 56.5 & 52.1 & 54.5 & 49.6 \\
Durall\cite{Durall}                        & 55.9 & 58.0 & 54.9 & 52.3 & 48.9 & 46.9 & 51.7 & 49.9 & 40.6 & 42.3 & 62.0 & 62.6 & 61.7 & 61.7 & 58.4 & 58.5 & 54.3 & 54.0 \\
Patchfor\cite{chai2020makes}                             & 79.8 & 99.1 & 87.3 & 99.7 & 82.8 & 99.1 & 84.9 & 98.8 & 74.2 & 81.4 & 95.8 & 99.8 & 95.6 & 99.9 & 94.0 & 99.8 & 86.8 & 97.2 \\
F3Net\cite{qian2020thinking}      & 71.6 & 79.9 & 88.3 & 95.4 & 87.0 & 94.5 & 88.5 & 95.4 & 69.2 & 70.8 & 74.1 & 84.0 & 73.4 & 83.3 & 80.7 & 89.1 & 79.1 & 86.5 \\
SelfBland\cite{shiohara2022detecting}                          & 52.4 & 51.6 & 58.8 & 63.2 & 59.4 & 64.1 & 64.2 & 68.3 & 58.3 & 63.4 & 53.0 & 54.0 & 52.6 & 51.9 & 51.9 & 52.6 & 56.3 & 58.7 \\
GANDetection\cite{mandelli2022detecting}                   & 67.2 & 83.0 & 51.2 & 52.6 & 51.1 & 51.9 & 51.7 & 53.5 & 49.6 & 49.0 & 54.7 & 65.8 & 54.9 & 65.9 & 53.8 & 58.9 & 54.3 & 60.1 \\
LGrad\cite{Tan2023CVPR}        & 88.5 & 97.3 & 89.4 & 94.9 & 87.4 & 93.2 & 90.7 & 95.1 & 86.6 & 100.0& 94.8 & 99.2 & 94.2 & 99.1 & 95.9 & 99.2 & \cblue{90.9} & \cblue{97.2} \\
UnivFD\cite{ojha2023towards}       & 89.5 & 96.8 & 90.1 & 97.0 & 90.7 & 97.2 & 91.1 & 97.4 & 75.7 & 85.1 & 90.5 & 97.0 & 90.2 & 97.1 & 77.3 & 88.6 & 86.9 & 94.5 \\
 \rowcolor{light-gray}MDIO (our)    &   93.2 & 98.0 & 98.9 & 99.6 & 98.8 & 99.5 & 98.8 & 99.6 & 82.2 & 83.8 & 98.7 & 99.5 & 98.8 & 99.5 & 98.6 & 99.5 & \cred{96.0} & \cred{97.4} \\

\bottomrule
    \end{tabular}
    }

  \label{tab:cvprdataset}
        \vspace{-0.25 cm}
\end{table*}

\begin{table*}[!ht]
    \centering
    \caption{ Evaluation on the Deepfake detection dataset of RPTC\cite{rptc}. The results of baselines   \cite{wang2020cnn,Frank,ju2022fusing,liu2020global,liu2022detecting,Tan2023CVPR,wang2023dire,ojha2023towards,rptc} presented in this table are obtained from \cite{rptc}. It should be noted that these baselines were trained using a 20-class configuration. In contrast, our proposed method adopts a more efficient 4-class configuration and utilizes significantly less training data for improved performance.}

     \resizebox{\textwidth}{31mm}{
    \begin{tabular}{l c c c c c c c c c c >{\columncolor{light-gray}}c}
    \bottomrule\hline
        Generator & CNNDet\cite{wang2020cnn} & FreDect\cite{Frank} & Fusing\cite{ju2022fusing} & GramNet\cite{liu2020global} & LNP\cite{liu2022detecting} & LGrad\cite{Tan2023CVPR} & DIRE-G\cite{wang2023dire} & DIRE-D\cite{wang2023dire} & UnivFD\cite{ojha2023towards} & RPTC\cite{rptc} & MDIO \\ \bottomrule\hline
        ProGAN & 100.00 & 99.36 & 100.00 & 99.99 & 99.67 & 99.83 & 95.19 & 52.75 & 99.81   & 100.00& 99.70 \\ 
        StyleGan & 90.17 & 78.02 & 85.20 & 87.05 & 91.75 & 91.08 & 83.03 & 51.31 & 84.93   & 92.77 & 99.52\\ 
        BigGAN & 71.17 & 81.97 & 77.40 & 67.33 & 77.75 & 85.62 & 70.12 & 49.70 & 95.08     & 95.80 & 84.80\\ 
        CycleGAN & 87.62 & 78.77 & 87.00 & 86.07 & 84.10 & 86.94 & 74.19 & 49.58 & 98.33   & 70.17 & 94.36\\ 
        StarGAN & 94.60 & 94.62 & 97.00 & 95.05 & 99.92 & 99.27 & 95.47 & 46.72 & 95.75    & 99.97 & 98.12\\ 
        GauGAN & 81.42 & 80.57 & 77.00 & 69.35 & 75.39 & 78.46 & 67.79 & 51.23 & 99.47     & 71.58 & 81.64\\ 
        Stylegan2 & 86.91 & 66.19 & 83.30 & 87.28 & 94.64 & 85.32 & 75.31 & 51.72 & 74.96  & 89.55 & 99.44\\ 
        WFIR & 91.65 & 50.75 & 66.80 & 86.80 & 70.85 & 55.70 & 58.05 & 53.30 & 86.90       & 85.80 & 57.05\\ 
        ADM & 60.39 & 63.42 & 49.00 & 58.61 & 84.73 & 67.15 & 75.78 & 98.25 & 66.87        & 82.17 & 95.24\\ 
        Glide & 58.07 & 54.13 & 57.20 & 54.50 & 80.52 & 66.11 & 71.75 & 92.42 & 62.46      & 83.79 & 97.56\\
        Midjourney & 51.39 & 45.87 & 52.20 & 50.02 & 65.55 & 65.35 & 58.01 & 89.45 & 56.13 & 90.12 & 96.78\\
        SDv1.4 & 50.57 & 38.79 & 51.00 & 51.70 & 85.55 & 63.02 & 49.74 & 91.24 & 63.66     & 95.38 & 93.76\\
        SDv1.5 & 50.53 & 39.21 & 51.40 & 52.16 & 85.67 & 63.67 & 49.83 & 91.63 & 63.49     & 95.30 & 93.67\\
        VQDM & 56.46 & 77.80 & 55.10 & 52.86 & 74.46 & 72.99 & 53.68 & 91.90 & 85.31       & 88.91 & 94.40\\
        Wukong & 51.03 & 40.30 & 51.70 & 50.76 & 82.06 & 59.55 & 54.46 & 90.90 & 70.93     & 91.07 & 93.06\\
        DALLE2 & 50.45 & 34.70 & 52.80 & 49.25 & 88.75 & 65.45 & 66.48 & 92.45 & 50.75     & 96.60 & 96.30\\ \bottomrule\hline
        Average & 70.78 & 64.03 & 68.38 & 68.67 & 83.84 & 75.34 & 68.68 & 71.53 & 78.43    & \ccb{89.31} & \cred{92.21}\\ \bottomrule\hline
    \end{tabular}
    }
    \label{tab:rptcdataset}
\end{table*}

\subsection{Results of Different Data-Independent Operators}
\label{DDIO}
In our Data-Independent Operator (DIO) framework, integrating the data-independent operator and a commonly used classification model results in significant improvements. 

\subsubsection{Handcrafted Filters and Pre-trained CNN Layers.} To assess the necessity of the data-independent operator, we conduct detection experiments both with and without its inclusion, using the ResNet50 model as the detector. 
In the absence of the data-independent operator, we train the detector using images directly.  
Additionally, we investigate the effectiveness of various filters as data-independent operators in extracting artifact representations, including Laplacian of Gaussian(LoG), Laplacian, Avgpool, Sobel filters, and the first layer of VGG16\cite{2014Very}, Resnet101\cite{he2016deep}, InceptionV3\cite{szegedy2016rethinking}, CLIP-Resnet50 \cite{CLIP}, and DeeplabV3\cite{DeeplabV3}. Instead of using the entire pretrained model as done in UnivFD\cite{ojha2023towards}, we only adopt the first layer. 

Specifically, we employ the 4 classes of ForenSynths \nocite{wang2020cnn} training set to train detectors, including car, cat, chair, and horse, generated by ProGAN, following baselines\cite{jeong2022bihpf,jeong2022frepgan,Tan2023CVPR}. 
Table \ref{tab:tab1} reports the results of Resnet50 with and without data-independent operators. 
We can observe that DIO consistently outperforms the classifier without the data-independent operator, across various fixed filters. 
 In contrast, when using the images as the artifact representation, the detector achieves high accuracy on the seen source ProGAN, but struggles to generalize to unseen sources, such as StyleGAN2 and BigGAN. 
However, by applying the first layer of Resnet101 as the data-independent operator, the performance on StyleGAN2 and BigGAN improves to 93.1\% and 76.9\%, respectively. 
The data-independent operator allows detectors to learn on high-quality artifact representations, leading to better performance on unknown sources. 
In particular, DIO-Resnet101 achieves the largest gain, rising by 8.1\% from 76.2\% to 84.1\% in terms of mean accuracy. 
Moreover, even with the Sobel filter consisting of only two \(3\times3\) matrices,  DIO-Sobel still achieves significant improvement, increasing the mean accuracy by 5.8\%. 
These results demonstrate the effectiveness of the data-independent operator consisting of handcrafted filters and pre-trained convolution layers in extracting the general artifact representation.

\subsubsection{Randomly Initialized Convolutional Layer}
We have investigated the suitability of various handcrafted filters and pre-trained convolution layers as data-independent operators to extract artifact representations. 
It is unclear whether prior knowledge is essential for the data-independent operator. 
According to \cite{ulyanov2018deep}, a randomly-initialized neural network can serve as an effective handcrafted prior.
Building on this idea, we further study the effectiveness of random filters in extracting the source-invariant representation. 
We employ randomly initialized convolutional layers with different shapes as the data-independent operator, as reported in Table \ref{tab:tab2}. 
It can be observed that layers with kernel size of \(1 \times 1\) surpass the learned and handcrafted filters. 
Among the various kernel shapes,  the (3, 64, 1, 1) kernel achieves the best performance, increasing the value of mean accuracy to 86.4\%. Compared with Resnet50 without the data-independent operator, DIO-Random exhibits a gain of 10.2\% in terms of mean accuracy.  
Moreover, DIO-Randoms outperform the handcrafted filters and pre-trained layers, particularly on unseen sources when the kernel size is \(1 \times 1\).
This suggests that prior knowledge is unnecessary for the data-independent operator. 

\subsubsection{The Fusion Structures of MDIO}
To further explore the potential of the data-independent operator (DIO), we investigate the use of multiple data-independent operators (MDIO) to enhance the generalization capabilities of detectors. 
We study the effects of different structures of MDIO on performance, using two data-independent operators - Sobel filter and randomly initialized layer with a kernel shape of (3, 64, 1, 1). Here we set the random seed as 88.
We test three fusion methods: early fusion, late fusion, and cascade fusion, achieving 85.2\%, 86.9\%, and 93.5\% on the ForenSynths test set in terms of mean Acc., respectively. 
We find that cascade fusion
 performs significantly better than early and late fusion, with an improvement of about 6\% to 8\%. 
The reason behind this is that cascade fusion applies multiple data-independent operators sequentially, allowing for more comprehensive filtering of the image content and mapping of the artifacts to more general representations. 

\subsection{Compared with SOTA }
To show the generalization ability of our DIO on unseen sources, we perform the evaluation on 5 deepfake detection datasets consisting of images from 33 generation models, e.g. ForenSynths\cite{wang2020cnn}, GANGen-Detection\cite{chuangchuangtan-GANGen-Detection}, DiffusionForensics\cite{Wang_2023_ICCV}, UniversalFakeDetect\cite{ojha2023towards}, AIGCDetectBenchmark\cite{rptc}. 
We compare with the previous methods: CNNDet\cite{wang2020cnn},  FreDect\cite{Frank}, Durall\cite{Durall} ,Patchfor\cite{chai2020makes},  F3Net\cite{qian2020thinking}, SelfBland\cite{shiohara2022detecting}, GANDetection\cite{mandelli2022detecting}, LGrad\cite{Tan2023CVPR}, UnivFD\cite{ojha2023towards}. Note that all the test results of MDIO in this article are trained with 4-classes setting ProGAN-$(car, cat, chair, horse)$.


\begin{figure*}[t]
  \centering
   \includegraphics[scale=0.57]{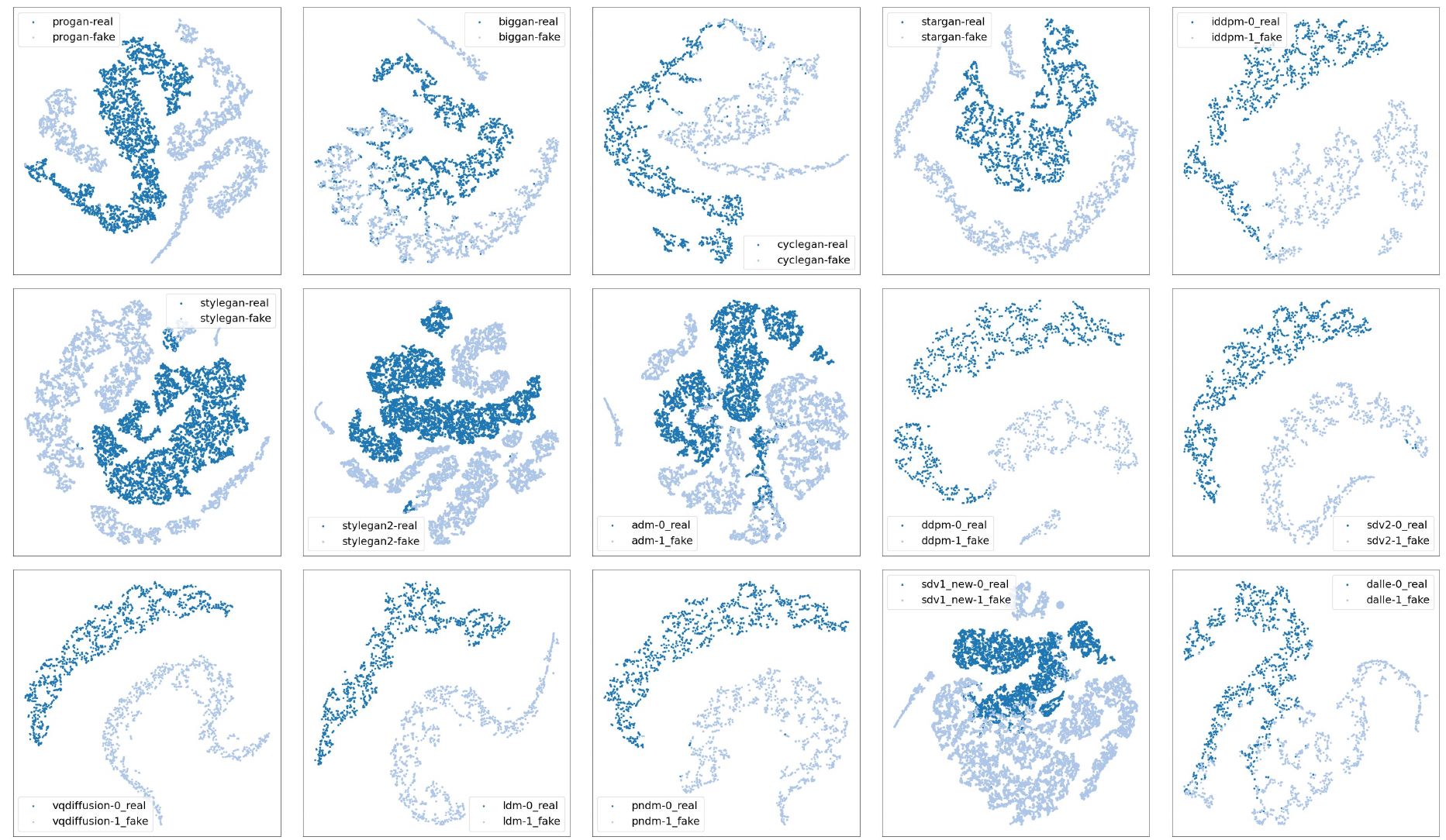}
   \caption{The t-SNE \cite{van2014accelerating} visualization of features extracted from the classifier. The blue and light blue point respect the feature of real and fake images, respectively.
   In the visualization, the fake images generated from 15 sources. It can be observed that the classifier trained on the ProGAN have ability to detect the cross-sources images, even diffusions. 
   This observation explains that Our proposed DIO successfully reduces the distribution drift between different sources, thereby enhancing the generalization ability of the detectors.
   }
   \label{fig:tsne}
\end{figure*}

\subsubsection{Evaluation on GAN models}
First, we evaluate our proposed methods on the GAN models, when they are trained on the ProGAN models. 
There are tow GAN test datesets, e.g. ForenSynths\cite{wang2020cnn} and GANGen-Detection\cite{chuangchuangtan-GANGen-Detection}, consisting of 16 GANs and Deepfake, in which 16 generation models is unseen in the training stage. 
In addition, the training set includes 4-classes of ProGAN, while the test set have 20-classes of ProGAN. 

The results of the ForenSynths are reported in Table \ref{tab:SOTA_ForenSynths}. 
The MDIO surpasses its counterparts in terms of mean Acc. and achieves comparable mean A.P. metric. 
Notably, with the 4-class setting, the mean Acc. values of DIO and MDIO are 86.4\% and 93.2\%, respectively.
Our MDIO outperforms the current state-of-the-art LGrad\cite{Tan2023CVPR} and UnivFD \cite{ojha2023towards} by 7.1\% and 4.1\% in terms of mean Acc. metric, while using fewer parameters. 
Our method adopt the Resnet50 as the backbone with 23.8M parameters, while the UnivFD leverages large pre-trained model CLIP with 304M parameters to extract the feature map. Although our method uses fewer parameters, it still achieves a 4.1\% improvement in accuracy. This can be attributed to the extraction of more general artifact features by {DIO}. 
To demonstrate the architectural independence of our method, we adopt Xception and Resnet50 with the removal of \(conv3\_x, conv4\_x,\) and \(conv5\_x\) layers as the backbone, and achieve mean accuracy of 91.2\% and 93.7\% on ForenSynths test set, respectively, surpassing all previous methods. 

We also show the results on GANGen-Detection\cite{chuangchuangtan-GANGen-Detection} in Table \ref{tab:SOTA_GANGen}. 
There are nine GAN models employed to assess the generalization capability of DIO. These models are trained on CelabA and ImageNet datasets.
It can be observed that, with the 4-class training setting of ProGAN, the proposed DIO and MDIO achieve the mean Acc. of 89.2\% and 95.2\% on those unseen GAN models, respectively. 
The MIO outperforms UnivFD \cite{ojha2023towards} by 6.0\% in terms of mean Acc. metric, showing  the efficacy of artifact representation of DIO. 


Furthermore, compared to CNNDet\cite{wang2020cnn}, our approach utilizes unlearnable filters exclusively prior to the classifier for the purpose of suppressing image content and converting it into a high-dimensional feature space, leading to notable enhancements. Specifically, we achieve an improvement of 26.1\% and 32.9\% in terms of mean accuracy on the ForenSynths dataset and GANGen-Detection dataset, respectively.


\subsubsection{Evaluation on Diffusion models}
To establish a more robust evaluation framework, we propose conducting an extensive experiment where the object detection model is trained using images synthesized by ProGAN, followed by testing on images generated through various diffusion models.
Two distinct datasets, namely DiffusionForensics and UniversalFakeDetect, are employed to evaluate the sources of diffusion-based image generation techniques. It should be noted that our approach leverages the ProGAN 4-classes configuration to maintain uniformity.
This evaluation framework aims to assess the efficacy of the object detection model when confronted with challenges posed by diffusion-based image generation techniques, thereby offering valuable insights into the generalization capability of Data-Independent Operator (DIO) across diverse image synthesis methodologies.

The performance evaluation results of DiffusionForensics \cite{Wang_2023_ICCV} are presented in Table \ref{tab:iccvdataset}. Despite being trained using images generated by ProGAN, MDIO exhibits strong generalization capabilities across different diffusion models. Our method achieves an average accuracy (Acc) of 98.7\% and an average precision (AP) of 99.8\%. In terms of Acc, MDIO surpasses existing state-of-the-art methods such as LGrad\cite{Tan2023CVPR} and UnivFD \cite{ojha2023towards} by margins of 10.5\% and 24.3\%, respectively.
Furthermore, when compared to DIRE \cite{Wang_2023_ICCV}, which specifically focuses on diffusion domain detection, MDIO demonstrates comparable results. This is particularly noteworthy considering that our training set consists of ProGAN images, while DIRE relies on diffusion models for training.

Given that a significant number of images in DiffusionForensics are classified as belonging to the bedroom class, we further evaluate its performance using the diffusion dataset provided by UniversalFakeDetect\cite{ojha2023towards}. In this dataset, \cite{ojha2023towards} utilizes diffusion models to generate images with either 100 or 200 iterations/steps. We report the results obtained on this diffusion dataset from UniversalFakeDetect in Table \ref{tab:cvprdataset}. Our proposed MDIO achieves an average accuracy score of 96\%, demonstrating its robustness across different scenarios. When compared against state-of-the-art methods such as LGrad and UnivFD, our MDIO surpasses them by margins of 5.1\% and 9.1\%, respectively, in terms of mean accuracy scores, highlighting its superior generalization capabilities across unseen diffusion datasets.



\subsubsection{Evaluation on AIGCDetectBenchmark}

Recently, a benchmark for deepfake detection called AIGCDetectBenchmark\cite{rptc} has been proposed. This benchmark includes the implementation of eight baseline methods as well as the design of a detector named RPTC using ProGAN with a 20-class training setting. Additionally, a comprehensive dataset is collected for this study. AIGCDetectBenchmark comprises sixteen different generative models, including GANs and Diffusions.
To demonstrate the generalizability of our DIO method and MDIO method, we conduct detection evaluations on the AIGCDetectBenchmark dataset. Table\ref{tab:rptcdataset} shows the results comparing MDIO with the baselines in terms of mean accuracy. It can be observed that our MDIO outperforms all baselines by a significant margin in terms of mean accuracy.


Comparison with RPTC\cite{rptc} shows that our proposed MDIO achieves a gain of 2.9\%.
By applying Data-Independent Operators on top of ResNet50, our method demonstrates a significant improvement of 21.42\% compared to CNNDet.
Furthermore, when DIRE utilizes diffusion images as training data, our MDIO trained on ProGAN dataset outperforms both DIRE-G by 23.52\% and DIRE-D by 20.67\%.
Despite being specifically trained using artificially generated images from the ProGAN model, our proposed method exhibits remarkable performance across various synthetic images, including diffusion-based ones.

\subsubsection{Parameters}

We show the number of parameters and mean Acc. of all 33 models in Table \ref{tab:Param}. Our MDIO outperforms UnivFD \cite{ojha2023towards}  13.3\% in terms of mean Acc. metric with less than one-tenth the number of parameters. 
The experimental results demonstrate the ability of the proposed DIO and MDIO to generalize well to different synthetic models, despite being trained only on images from ProGANs. 
The results confirm the generalization capability of the proposed data-independent operator to extract a general representation of artifacts, and generalize this representation across various synthetic models and categories.

\begin{table}[!ht]
    \centering  \caption{{Parameters and mean Results on all test set..} }
\resizebox{58mm}{12.3mm}{
    \begin{tabular}{c | c c}
    \bottomrule \hline
Methods & Parameters & mean Acc. \\ \bottomrule \hline
UnivFD\cite{wang2020cnn}& 23.51  & 62.5 \\ \hline 
FreDect\cite{Frank}             & 23.51  & 58.9 \\ \hline 
LGrad \cite{Tan2023CVPR}      & 46.56  & 80.5 \\ \hline 
UnivFD \cite{ojha2023towards}   & 332.32 & 80.7 \\ \hline 
MDIO                          & 23.70 & \cred{94.6}
 \\ \hline 
  \end{tabular} }
  \label{tab:Param}
\end{table}

\begin{table}[!ht]
    \centering  \caption{{Effect of Random seed. Results on ForenSynths test set.} }
\resizebox{58mm}{11.3mm}{
    \begin{tabular}{c | c c}
    \bottomrule \hline
     Random seed & mean Acc. & mean A.P. \\ \bottomrule \hline

9              &   91.9 & 95.5 \\ \hline  
88              & 93.2  & 95.1  \\ \hline 
321              & 91.1  & 96.8   \\ \hline 
888           & 91.2  & 97.4   \\ \bottomrule 

    \end{tabular} }

  \label{tab:seed}
\end{table}
\begin{table}[!ht]
    \centering  \caption{{Effect of Backbone. Results on ForenSynths test set.} }
\resizebox{38mm}{9.7mm}{
    \begin{tabular}{c  c }
    \bottomrule \hline
Backbones & mean Acc. \\ \bottomrule \hline
Resnet50\_-plain &   91.2  \\ \hline 
Resnet50         &  93.2 \\ \hline 
Xception         & 93.7    \\ \hline 
  \end{tabular} }
  \label{tab:seed}
\end{table}

\subsection{Ablation study}
\subsubsection{Effect of Random seed}
In order to comprehensively assess the impact of the randomly initialized convolutional layers within MDIO, we set different random seeds during training.  Subsequently, we evaluate the MDIO's performance across the ForenSynths \cite{wang2020cnn} test set. 
The results are tabulated in Table \ref{tab:seed}. The proposed MDIO consistently achieves noteworthy results across different random seeds, attaining mean accuracy metrics of 91.9\%, 93.5\%, 91.1\%, and 91.2\% with 9, 88, 321, and 888 random seeds, respectively. All  results surpass the current state-of-the-art LGrad \cite{Tan2023CVPR} and UnivFD \cite{ojha2023towards}. This evidence reinforces the effectiveness of our MDIO framework.

\subsubsection{Backbone}
To demonstrate the architectural independence of our method, we adopt Xception and Resnet50 with the removal of \(conv3\_x, conv4\_x,\) and \(conv5\_x\) layers as the backbone, and achieve mean accuracy of 91.2\% and 93.7\% on ForenSynths test set, respectively, surpassing all previous methods.

\subsubsection{Visualization}

\textbf{t-SNE.} 
To qualitatively verify the effectiveness of the proposed DIO, we employ t-SNE \cite{van2014accelerating} visualization of classifier feature from 15 sources in the Fig. \ref{fig:tsne}, demonstrating the DIO adopt the source-invariant artifact representations from the images of ProGAN. 
We can observe that although the classifier is trained on ProGAN, it can still effectively detect images from different sources. 



\section{Conclusion}
This work devotes attention to developing more generic and effective artifact representation for generalizable forgery image detection. 
To achieve this goal, we introduce the data-independent operator (DIO) framework, a simple yet effective method that outperforms existing state-of-the-art approaches. 
To develop source-invariant artifact representations, our DIO framework employs filters that remain independent of the training source as data-independent operators, which are fixed during the training stage. 
In addition, we build upon DIO and propose Multiple DIO to further improve the performance. 
Extensive experiments on 33 generation models demonstrate the generalization ability of the proposed DIO in extracting general artifact representations. Our detector achieves a new SOTA performance, with a remarkable gain of 13.3\%.  Our DIO and MDIO can serve as strong baselines for future studies.

\bibliographystyle{IEEEtran}

\bibliography{egbib.bib}

\begin{thebibliography}{10}
\providecommand{\url}[1]{#1}
\csname url@samestyle\endcsname
\providecommand{\newblock}{\relax}
\providecommand{\bibinfo}[2]{#2}
\providecommand{\BIBentrySTDinterwordspacing}{\spaceskip=0pt\relax}
\providecommand{\BIBentryALTinterwordstretchfactor}{4}
\providecommand{\BIBentryALTinterwordspacing}{\spaceskip=\fontdimen2\font plus
\BIBentryALTinterwordstretchfactor\fontdimen3\font minus \fontdimen4\font\relax}
\providecommand{\BIBforeignlanguage}[2]{{%
\expandafter\ifx\csname l@#1\endcsname\relax
\typeout{** WARNING: IEEEtran.bst: No hyphenation pattern has been}%
\typeout{** loaded for the language `#1'. Using the pattern for}%
\typeout{** the default language instead.}%
\else
\language=\csname l@#1\endcsname
\fi
#2}}
\providecommand{\BIBdecl}{\relax}
\BIBdecl

\bibitem{goodfellow2014generative}
I.~J. Goodfellow \emph{et~al.}, ``Generative adversarial nets,'' in \emph{NIPS}, 2014.

\bibitem{karras2018progressive}
T.~Karras \emph{et~al.}, ``Progressive growing of gans for improved quality, stability, and variation,'' in \emph{International Conference on Learning Representations}, 2018.

\bibitem{karras2019style}
------, ``A style-based generator architecture for generative adversarial networks,'' in \emph{Proceedings of the CVPR}, 2019, pp. 4401--4410.

\bibitem{ho2020denoising}
J.~Ho \emph{et~al.}, ``Denoising diffusion probabilistic models,'' \emph{Advances in neural information processing systems}, vol.~33, pp. 6840--6851, 2020.

\bibitem{rombach2022high}
R.~Rombach \emph{et~al.}, ``High-resolution image synthesis with latent diffusion models,'' in \emph{Proceedings of the IEEE/CVF conference on computer vision and pattern recognition}, 2022, pp. 10\,684--10\,695.

\bibitem{wang2020cnn}
S.-Y. Wang \emph{et~al.}, ``Cnn-generated images are surprisingly easy to spot... for now,'' in \emph{Proceedings of the CVPR}, 2020, pp. 8695--8704.

\bibitem{Frank}
J.~Frank \emph{et~al.}, ``Leveraging frequency analysis for deep fake image recognition,'' in \emph{International conference on machine learning}.\hskip 1em plus 0.5em minus 0.4em\relax PMLR, 2020, pp. 3247--3258.

\bibitem{frank2020leveraging}
------, ``Leveraging frequency analysis for deep fake image recognition,'' in \emph{International conference on machine learning}.\hskip 1em plus 0.5em minus 0.4em\relax PMLR, 2020, pp. 3247--3258.

\bibitem{jeong2022frepgan}
Y.~Jeong \emph{et~al.}, ``Frepgan: robust deepfake detection using frequency-level perturbations,'' in \emph{Proceedings of the AAAI Conference on Artificial Intelligence}, vol.~36, no.~1, 2022, pp. 1060--1068.

\bibitem{jeong2022fingerprintnet}
------, ``Fingerprintnet: Synthesized fingerprints for generated image detection,'' in \emph{European Conference on Computer Vision}.\hskip 1em plus 0.5em minus 0.4em\relax Springer, 2022, pp. 76--94.

\bibitem{chen2022ost}
L.~Chen \emph{et~al.}, ``Ost: Improving generalization of deepfake detection via one-shot test-time training,'' in \emph{Advances in Neural Information Processing Systems}, 2022.

\bibitem{ojha2023towards}
U.~Ojha \emph{et~al.}, ``Towards universal fake image detectors that generalize across generative models,'' in \emph{Proceedings of the CVPR}, 2023, pp. 24\,480--24\,489.

\bibitem{Tan2023CVPR}
C.~Tan \emph{et~al.}, ``Learning on gradients: Generalized artifacts representation for gan-generated images detection,'' in \emph{Proceedings of the CVPR (CVPR)}, June 2023, pp. 12\,105--12\,114.

\bibitem{ulyanov2018deep}
D.~Ulyanov \emph{et~al.}, ``Deep image prior,'' in \emph{Proceedings of the IEEE conference on computer vision and pattern recognition}, 2018, pp. 9446--9454.

\bibitem{Deepfake}
A.~Rossler \emph{et~al.}, ``Faceforensics++: Learning to detect manipulated facial images,'' in \emph{Proceedings of the ICCV}, 2019, pp. 1--11.

\bibitem{chollet2017xception}
F.~Chollet, ``Xception: Deep learning with depthwise separable convolutions,'' in \emph{Proceedings of the IEEE conference on computer vision and pattern recognition}, 2017, pp. 1251--1258.

\bibitem{li2018ictu}
Y.~Li \emph{et~al.}, ``In ictu oculi: Exposing ai created fake videos by detecting eye blinking,'' in \emph{2018 IEEE International workshop on information forensics and security (WIFS)}.\hskip 1em plus 0.5em minus 0.4em\relax IEEE, 2018, pp. 1--7.

\bibitem{haliassos2021lips}
A.~Haliassos \emph{et~al.}, ``Lips don't lie: A generalisable and robust approach to face forgery detection,'' in \emph{Proceedings of the CVPR}, 2021, pp. 5039--5049.

\bibitem{ye2007detecting}
S.~Ye \emph{et~al.}, ``Detecting digital image forgeries by measuring inconsistencies of blocking artifact,'' in \emph{2007 IEEE ICME}.\hskip 1em plus 0.5em minus 0.4em\relax Ieee, 2007, pp. 12--15.

\bibitem{zhao2021learning}
T.~Zhao \emph{et~al.}, ``Learning self-consistency for deepfake detection,'' in \emph{Proceedings of the ICCV}, 2021, pp. 15\,023--15\,033.

\bibitem{dong2022protecting}
X.~Dong \emph{et~al.}, ``Protecting celebrities from deepfake with identity consistency transformer,'' in \emph{Proceedings of the CVPR}, 2022, pp. 9468--9478.

\bibitem{chai2020makes}
L.~Chai \emph{et~al.}, ``What makes fake images detectable? understanding properties that generalize,'' in \emph{European conference on computer vision}.\hskip 1em plus 0.5em minus 0.4em\relax Springer, 2020, pp. 103--120.

\bibitem{wang2021representative}
C.~Wang \emph{et~al.}, ``Representative forgery mining for fake face detection,'' in \emph{Proceedings of the CVPR}, 2021, pp. 14\,923--14\,932.

\bibitem{chen2022self}
L.~Chen \emph{et~al.}, ``Self-supervised learning of adversarial example: Towards good generalizations for deepfake detection,'' in \emph{Proceedings of the CVPR}, 2022, pp. 18\,710--18\,719.

\bibitem{cao2022end}
J.~Cao \emph{et~al.}, ``End-to-end reconstruction-classification learning for face forgery detection,'' in \emph{Proceedings of the CVPR}, 2022, pp. 4113--4122.

\bibitem{he2021beyond}
Y.~He \emph{et~al.}, ``Beyond the spectrum: Detecting deepfakes via re-synthesis,'' in \emph{Proceedings of the Thirtieth International Joint Conference on Artificial Intelligence, IJCAI-21}.\hskip 1em plus 0.5em minus 0.4em\relax International Joint Conferences on Artificial Intelligence Organization, 2021, pp. 2534--2541.

\bibitem{shiohara2022detecting}
K.~Shiohara \emph{et~al.}, ``Detecting deepfakes with self-blended images,'' in \emph{Proceedings of the CVPR}, 2022, pp. 18\,720--18\,729.

\bibitem{ju2022fusing}
Y.~Ju \emph{et~al.}, ``Fusing global and local features for generalized ai-synthesized image detection,'' in \emph{2022 IEEE International Conference on Image Processing (ICIP)}.\hskip 1em plus 0.5em minus 0.4em\relax IEEE, 2022, pp. 3465--3469.

\bibitem{Durall}
R.~Durall \emph{et~al.}, ``Watch your up-convolution: Cnn based generative deep neural networks are failing to reproduce spectral distributions,'' in \emph{Proceedings of the CVPR}, 2020, pp. 7890--7899.

\bibitem{masi2020two}
I.~Masi \emph{et~al.}, ``Two-branch recurrent network for isolating deepfakes in videos,'' in \emph{European conference on computer vision}.\hskip 1em plus 0.5em minus 0.4em\relax Springer, 2020, pp. 667--684.

\bibitem{qian2020thinking}
Y.~Qian \emph{et~al.}, ``Thinking in frequency: Face forgery detection by mining frequency-aware clues,'' in \emph{European conference on computer vision}.\hskip 1em plus 0.5em minus 0.4em\relax Springer, 2020, pp. 86--103.

\bibitem{luo2021generalizing}
Y.~Luo \emph{et~al.}, ``Generalizing face forgery detection with high-frequency features,'' in \emph{Proceedings of the CVPR}, 2021, pp. 16\,317--16\,326.

\bibitem{woo2022add}
S.~Woo \emph{et~al.}, ``Add: Frequency attention and multi-view based knowledge distillation to detect low-quality compressed deepfake images,'' in \emph{Proceedings of the AAAI Conference on Artificial Intelligence}, vol.~36, no.~1, 2022, pp. 122--130.

\bibitem{jeong2022bihpf}
Y.~Jeong \emph{et~al.}, ``Bihpf: Bilateral high-pass filters for robust deepfake detection,'' in \emph{Proceedings of the IEEE/CVF Winter Conference on Applications of Computer Vision}, 2022, pp. 48--57.

\bibitem{karras2020analyzing}
T.~Karras \emph{et~al.}, ``Analyzing and improving the image quality of stylegan,'' in \emph{Proceedings of the CVPR}, 2020, pp. 8110--8119.

\bibitem{BigGAN}
A.~Brock \emph{et~al.}, ``Large scale gan training for high fidelity natural image synthesis,'' in \emph{International Conference on Learning Representations}, 2018.

\bibitem{CycleGAN}
J.-Y. Zhu \emph{et~al.}, ``Unpaired image-to-image translation using cycle-consistent adversarial networks,'' in \emph{Proceedings of the IEEE international conference on computer vision}, 2017, pp. 2223--2232.

\bibitem{choi2018stargan}
Y.~Choi \emph{et~al.}, ``Stargan: Unified generative adversarial networks for multi-domain image-to-image translation,'' in \emph{Proceedings of the IEEE conference on computer vision and pattern recognition}, 2018, pp. 8789--8797.

\bibitem{GauGAN}
T.~Park \emph{et~al.}, ``Semantic image synthesis with spatially-adaptive normalization,'' in \emph{Proceedings of the CVPR}, 2019, pp. 2337--2346.

\bibitem{yu2015lsun}
F.~Yu \emph{et~al.}, ``Lsun: Construction of a large-scale image dataset using deep learning with humans in the loop,'' \emph{arXiv preprint arXiv:1506.03365}, 2015.

\bibitem{russakovsky2015imagenet}
O.~Russakovsky \emph{et~al.}, ``Imagenet large scale visual recognition challenge,'' \emph{International journal of computer vision}, vol. 115, no.~3, pp. 211--252, 2015.

\bibitem{CelebA}
Z.~Liu \emph{et~al.}, ``Deep learning face attributes in the wild,'' in \emph{Proceedings of the IEEE international conference on computer vision}, 2015, pp. 3730--3738.

\bibitem{coco}
T.-Y. Lin \emph{et~al.}, ``Microsoft coco: Common objects in context,'' in \emph{European conference on computer vision}.\hskip 1em plus 0.5em minus 0.4em\relax Springer, 2014, pp. 740--755.

\bibitem{chuangchuangtan-GANGen-Detection}
H.~L. Chuangchuang~Tan, Renshuai~Tao, ``Gangen-detection: A dataset generated by gans for generalizable deepfake detection,'' \url{https://github.com/chuangchuangtan/GANGen-Detection}, 2024.

\bibitem{AttGAN}
Z.~He \emph{et~al.}, ``Attgan: Facial attribute editing by only changing what you want,'' \emph{IEEE Transactions on Image Processing}, vol.~28, no.~11, pp. 5464--5478, 2019.

\bibitem{began}
D.~Berthelot \emph{et~al.}, ``Began: Boundary equilibrium generative adversarial networks,'' \emph{arXiv preprint arXiv:1703.10717}, 2017.

\bibitem{CramerGAN}
M.~G. Bellemare \emph{et~al.}, ``The cramer distance as a solution to biased wasserstein gradients,'' \emph{arXiv preprint arXiv:1705.10743}, 2017.

\bibitem{InfoMaxGAN}
K.~S. Lee \emph{et~al.}, ``Infomax-gan: Improved adversarial image generation via information maximization and contrastive learning,'' in \emph{Proceedings of the IEEE/CVF winter conference on applications of computer vision}, 2021, pp. 3942--3952.

\bibitem{MMDGAN}
C.-L. Li \emph{et~al.}, ``Mmd gan: Towards deeper understanding of moment matching network,'' \emph{Advances in neural information processing systems}, vol.~30, 2017.

\bibitem{RelGAN}
W.~Nie \emph{et~al.}, ``Relgan: Relational generative adversarial networks for text generation,'' in \emph{International conference on learning representations}, 2019.

\bibitem{S3GAN}
M.~Lu{\v{c}}i{\'c} \emph{et~al.}, ``High-fidelity image generation with fewer labels,'' in \emph{International conference on machine learning}.\hskip 1em plus 0.5em minus 0.4em\relax PMLR, 2019, pp. 4183--4192.

\bibitem{SNGAN}
T.~Miyato \emph{et~al.}, ``Spectral normalization for generative adversarial networks,'' \emph{arXiv preprint arXiv:1802.05957}, 2018.

\bibitem{STGAN}
M.~Liu \emph{et~al.}, ``Stgan: A unified selective transfer network for arbitrary image attribute editing,'' in \emph{Proceedings of the CVPR}, 2019, pp. 3673--3682.

\bibitem{Wang_2023_ICCV}
Z.~Wang \emph{et~al.}, ``Dire for diffusion-generated image detection,'' in \emph{Proceedings of the IEEE/CVF International Conference on Computer Vision (ICCV)}, October 2023, pp. 22\,445--22\,455.

\bibitem{dhariwal2021diffusion}
P.~Dhariwal \emph{et~al.}, ``Diffusion models beat gans on image synthesis,'' \emph{Advances in neural information processing systems}, vol.~34, pp. 8780--8794, 2021.

\bibitem{nichol2021improved}
A.~Q. Nichol \emph{et~al.}, ``Improved denoising diffusion probabilistic models,'' in \emph{International Conference on Machine Learning}.\hskip 1em plus 0.5em minus 0.4em\relax PMLR, 2021, pp. 8162--8171.

\bibitem{liu2022pseudo}
L.~Liu \emph{et~al.}, ``Pseudo numerical methods for diffusion models on manifolds,'' in \emph{ICLR}, 2022.

\bibitem{gu2022vector}
S.~Gu \emph{et~al.}, ``Vector quantized diffusion model for text-to-image synthesis,'' in \emph{Proceedings of the IEEE/CVF Conference on Computer Vision and Pattern Recognition}, 2022, pp. 10\,696--10\,706.

\bibitem{nichol2021glide}
A.~Nichol \emph{et~al.}, ``Glide: Towards photorealistic image generation and editing with text-guided diffusion models,'' \emph{arXiv preprint arXiv:2112.10741}, 2021.

\bibitem{ramesh2021zero}
A.~Ramesh \emph{et~al.}, ``Zero-shot text-to-image generation,'' in \emph{ICML}.\hskip 1em plus 0.5em minus 0.4em\relax PMLR, 2021, pp. 8821--8831.

\bibitem{schuhmann2021laion}
C.~Schuhmann \emph{et~al.}, ``Laion-400m: Open dataset of clip-filtered 400 million image-text pairs,'' in \emph{NeurIPS Workshop Datacentric AI}, no. FZJ-2022-00923.\hskip 1em plus 0.5em minus 0.4em\relax J{\"u}lich Supercomputing Center, 2021.

\bibitem{rptc}
N.~Zhong, Y.~Xu, Z.~Qian, and X.~Zhang, ``Rich and poor texture contrast: A simple yet effective approach for ai-generated image detection,'' \emph{arXiv preprint arXiv:2311.12397}, 2023.

\bibitem{Midjourney}
``Midjourney,'' \url{https://www.midjourney.com/home/}, 2023.

\bibitem{Wukong}
``wukong,'' \url{https://xihe.mindspore.cn/modelzoo/wukong}, 2023.

\bibitem{zhu2023genimage}
M.~Zhu, H.~Chen, Q.~Yan, X.~Huang, G.~Lin, W.~Li, Z.~Tu, H.~Hu, J.~Hu, and Y.~Wang, ``Genimage: A million-scale benchmark for detecting ai-generated image,'' 2023.

\bibitem{ramesh2022hierarchical}
A.~Ramesh, P.~Dhariwal, A.~Nichol, C.~Chu, and M.~Chen, ``Hierarchical text-conditional image generation with clip latents,'' \emph{arXiv preprint arXiv:2204.06125}, vol.~1, no.~2, p.~3, 2022.

\bibitem{he2016deep}
K.~He \emph{et~al.}, ``Deep residual learning for image recognition,'' in \emph{Proceedings of the IEEE conference on computer vision and pattern recognition}, 2016, pp. 770--778.

\bibitem{kingma2015adam}
D.~P. Kingma and J.~Ba, ``Adam: A method for stochastic optimization,'' in \emph{ICLR (Poster)}, 2015.

\bibitem{2014Very}
K.~Simonyan \emph{et~al.}, ``Very deep convolutional networks for large-scale image recognition,'' \emph{International Conference on Learning Representations}, 2015.

\bibitem{szegedy2016rethinking}
C.~Szegedy \emph{et~al.}, ``Rethinking the inception architecture for computer vision,'' in \emph{Proceedings of the IEEE conference on computer vision and pattern recognition}, 2016, pp. 2818--2826.

\bibitem{CLIP}
A.~Radford \emph{et~al.}, ``Learning transferable visual models from natural language supervision,'' in \emph{International Conference on Machine Learning}.\hskip 1em plus 0.5em minus 0.4em\relax PMLR, 2021, pp. 8748--8763.

\bibitem{DeeplabV3}
L.-C. Chen \emph{et~al.}, ``Rethinking atrous convolution for semantic image segmentation,'' \emph{arXiv preprint arXiv:1706.05587}, 2017.

\bibitem{mandelli2022detecting}
S.~Mandelli, N.~Bonettini, P.~Bestagini, and S.~Tubaro, ``Detecting gan-generated images by orthogonal training of multiple cnns,'' in \emph{2022 IEEE International Conference on Image Processing (ICIP)}, 2022, pp. 3091--3095.

\bibitem{liu2020global}
Z.~Liu \emph{et~al.}, ``Global texture enhancement for fake face detection in the wild,'' in \emph{Proceedings of the CVPR}, 2020, pp. 8060--8069.

\bibitem{liu2022detecting}
B.~Liu, F.~Yang, X.~Bi, B.~Xiao, W.~Li, and X.~Gao, ``Detecting generated images by real images,'' in \emph{European Conference on Computer Vision}.\hskip 1em plus 0.5em minus 0.4em\relax Springer, 2022, pp. 95--110.

\bibitem{wang2023dire}
Z.~Wang, J.~Bao, W.~Zhou, W.~Wang, H.~Hu, H.~Chen, and H.~Li, ``Dire for diffusion-generated image detection,'' \emph{arXiv preprint arXiv:2303.09295}, 2023.

\bibitem{van2014accelerating}
L.~Van Der~Maaten, ``Accelerating t-sne using tree-based algorithms,'' \emph{The journal of machine learning research}, vol.~15, no.~1, pp. 3221--3245, 2014.

\end{thebibliography}


\newpage

 




\vfill

\end{document}